\documentclass[11pt]{article}
\usepackage[letterpaper,margin=1in]{geometry}
\usepackage{libertine}

\usepackage[utf8]{inputenc} 
\usepackage[T1]{fontenc}    
\usepackage[colorlinks]{hyperref}       
\usepackage{url}            
\usepackage{booktabs}       
\usepackage{amsfonts}       
\usepackage{nicefrac}       
\usepackage{microtype}      

\usepackage{colortbl}
\usepackage{xcolor}
\usepackage{enumitem}
\usepackage{multirow}

\usepackage{amsmath}
\usepackage{amssymb}
\usepackage{amsfonts}
\usepackage{mathtools}
\usepackage{bm}
\usepackage{bbm}
\usepackage{nicefrac}
\usepackage{centernot}

\usepackage{float}
\usepackage{graphicx}
\usepackage{wrapfig}
\usepackage{subcaption}
\graphicspath{{./figures/}}
\DeclareGraphicsExtensions{.pdf,.png,.jpeg,.jpg}

\usepackage{algorithm}
\usepackage{algorithmic}

\usepackage[capitalize,sort&compress]{cleveref}

\definecolor{blue}{rgb}{0,0.17647059,0.44705882}
\definecolor{green}{HTML}{008767}
\definecolor{red}{HTML}{CF4520}
\definecolor{orange}{HTML}{F56600}
\definecolor{purple}{HTML}{A45C98}

\usepackage[numbers,sort&compress]{natbib}

\definecolor{blue}{rgb}{0,0.17647059,0.44705882}
\definecolor{green}{HTML}{008767}
\definecolor{red}{HTML}{CF4520}
\definecolor{orange}{HTML}{F56600}
\definecolor{purple}{HTML}{A45C98}
\def\X{{\mathcal X}}
\def\Y{{\mathcal Y}}
\def\U{{\mathcal U}}

\def\W{{\mathcal W}}
\def\L{{\mathcal L}}
\def\D{{\mathcal D}}
\def\C{{\mathcal C}}

\def\G{{\mathcal G}}

\def\E{{\mathbb E}}
\def\P{{\mathbb P}}
\def\R{{\mathbb R}}

\def\N{{\mathbb N}}

\def\1{{\bm 1}}

\def\y{{\hat y}}
\def\z{{\hat z}}

\def\pref{{\pi_{\text{ref}}}}

\def\KL{{\text{KL}}}

\def\eg{{\text{e.g.}}}
\def\ie{{\text{i.e.}}}
\def\th{{^\text{th}}}
\def\rref{{\text{ref}}}
\def\pref{{\text{pref}}}
\def\expl{{\text{expl}}}

\def\cons{{\texttt{consitency}}}
\def\DPO{{\text{DPO}}}
\def\TP{{\text{TP}}}
\def\TN{{\text{TN}}}

\newcommand{\at}[1]{^{(#1)}}
\newcommand{\mline}[1]{\multicolumn{1}{c}{\begin{tabular}[c]{@{}c@{}}#1\end{tabular}}}
\newcommand{\acc}[1]{\color{gray!70}(#1)}

\DeclareMathOperator*{\argmax}{\arg\max}

\usepackage[numbers,sort&compress]{natbib}

\title{\textbf{Direct Preference Optimization for\\Adaptive Concept-based Explanations}}

\author{
    Jacopo~Teneggi$^1$
    \quad
    Zhenzhen~Wang$^1$
    \quad
    Paul~H.~Yi$^2$
    \quad
    Tianmin~Shu$^1$
    \quad
    Jeremias~Sulam$^1$\\
    \\
    $^1$Johns Hopkins University\\
    $^2$St. Jude Children's Research Hospital\\
    {\small\texttt{\{jtenegg1,zwang218,tianmin.shu,jsulam1\}@jhu.edu}}, {\small\texttt{paul.yi@stjude.org}}
}
\date{
    \centering
    \rule{0.05\linewidth}{.1pt}\\
    \begin{minipage}{0.6\linewidth}
        \centering
        \begin{align*}
        \text{code:}    &&~\text{\scriptsize\url{https://github.com/Sulam-Group/pragmatixs}}
        \end{align*}
    \end{minipage}
}

\begin{document}
\maketitle
\begin{abstract}
\noindent Concept-based explanation methods aim at making machine learning models more transparent by finding the most important semantic features of an input (\eg, colors, patterns, shapes) for a given prediction task.
However, these methods generally ignore the communicative context of explanations, such as the preferences of a listener. For example, medical doctors understand explanations in terms of clinical markers, but patients may not, needing a different vocabulary to rationalize the same diagnosis. 
We address this gap with listener-adaptive explanations grounded in principles of pragmatic reasoning and the rational speech act. We introduce an iterative training procedure based on direct preference optimization where a speaker learns to compose explanations that maximize communicative utility for a listener.
Our approach only needs access to pairwise preferences, which can be collected from human feedback, making it particularly relevant in real-world scenarios where a model of the listener may not be available.
We demonstrate that our method is able to align speakers with the preferences of simulated listeners on image classification across three datasets, and further validate that pragmatic explanations generated with our method improve the classification accuracy of participants in a user study.
\end{abstract}

\section{Introduction}
Understanding the decision-making process of modern machine learning systems is critical for their responsible use. This need has motivated considerable research efforts on the safety, fairness, and trustworthiness of black-box predictors. A notable approach is that of \emph{explaining} predictions in a post-hoc fashion, \ie~finding the features that contributed the most to the output of a model, either globally over a population or locally for a given input \cite{covert2020understanding}. Instead of using input features (\eg, pixels for images or words for text), recent works have proposed to compose explanations with human-interpretable concepts, such as the presence of certain objects, colors, or patterns in images \cite{kim2018interpretability,koh2020concept,yuksekgonul2022post,teneggi2024testing}. Even though these approaches are closer to the way we articulate explanations compared to input features, they overlook the \emph{context} of explanation (\ie, its \emph{pragmatics}), which is a fundamental aspect of reasoning \cite{grice1975logic,hobbs1987implicature,carston1993conjunction,wilson2012linguistic,recanati1989pragmatics,gibbs1997pragmatics}.

In conversation, we do not interpret sentences by their literal meaning only, but instead integrate them with contextual information. For example, we take into consideration our knowledge of the world and that of our interlocutors, greatly influencing the way we understand sentences from, or produce justifications to, other people. Therefore, it is natural to expect that machine learning explanations be pragmatic to be effective at communicating with the users of a model. These aspects can become crucial in scenarios with various stakeholders, where \emph{listeners}---those who receive an explanation---have considerably different experiences, knowledge, and intentions. In this way, a pragmatic explainer should reason about whom it is conversing with (\eg, a medical doctor, a nurse, or a patient), adapting to maximize communicative utility. These considerations on the pragmatic aspects of explanation have a long-standing tradition in philosophy \cite{bromberger1992we,bromberger1984pragmatic,de2017understanding,van1980scientific} and in cognitive sciences \cite{harding2025communication,goddu2024development,goodman2016pragmatic}, having found application primarily in computational linguistics \cite{krahmer2012computational,schlotterbeck2023incremental,cohn2018pragmatically,ou2023pragmatic}. While the importance of pragmatic reasoning to improve human-machine interaction in explainable artificial intelligence has been raised both conceptually \cite{nyrup2022explanatory,tsai2020logic,paez2019pragmatic,miller2019explanation,schneider2019personalized,kim2025because} and empirically \cite{zhang2020effect,ehsan2021expanding,kaur2020interpreting,panigrahi2025interactivity,bertrand2023selective,corti2024moving,haque2023explainable,schoonderwoerd2021human}, technical exploration of listener-adaptive explanation methods remains limited \cite{wu2024discret,wang2018reinforcement}.

In this work, we introduce a framework that tailors explanations to the needs of different listeners. In particular, following the rational speech act (RSA) \cite{goodman2016pragmatic}, we formalize pragmatic reasoning as probabilistic inference over explanations. We couple this formalism with direct preference optimization (DPO) \cite{rafailov2023direct} in order to encompass real-world scenarios where explicit listener models may be hard to estimate from data, but preferences can be collected from human feedback. For a fixed black-box classifier whose predictions we want to explain, we propose an iterative training procedure between a \emph{speaker}---who generates explanations---and a listener. Akin to cooperative signaling games \cite{sobel2020signaling}, the goal of the speaker is to compose concept-based explanations that maximize the ability of the listener to infer the prediction of the base classifier (\ie, the signal). Importantly, the listener does not have access to the input of the classifier, but only to the utterance of the speaker (\ie, the explanation).

\begin{figure*}[t]
    \centering
    \includegraphics[width=0.8\linewidth]{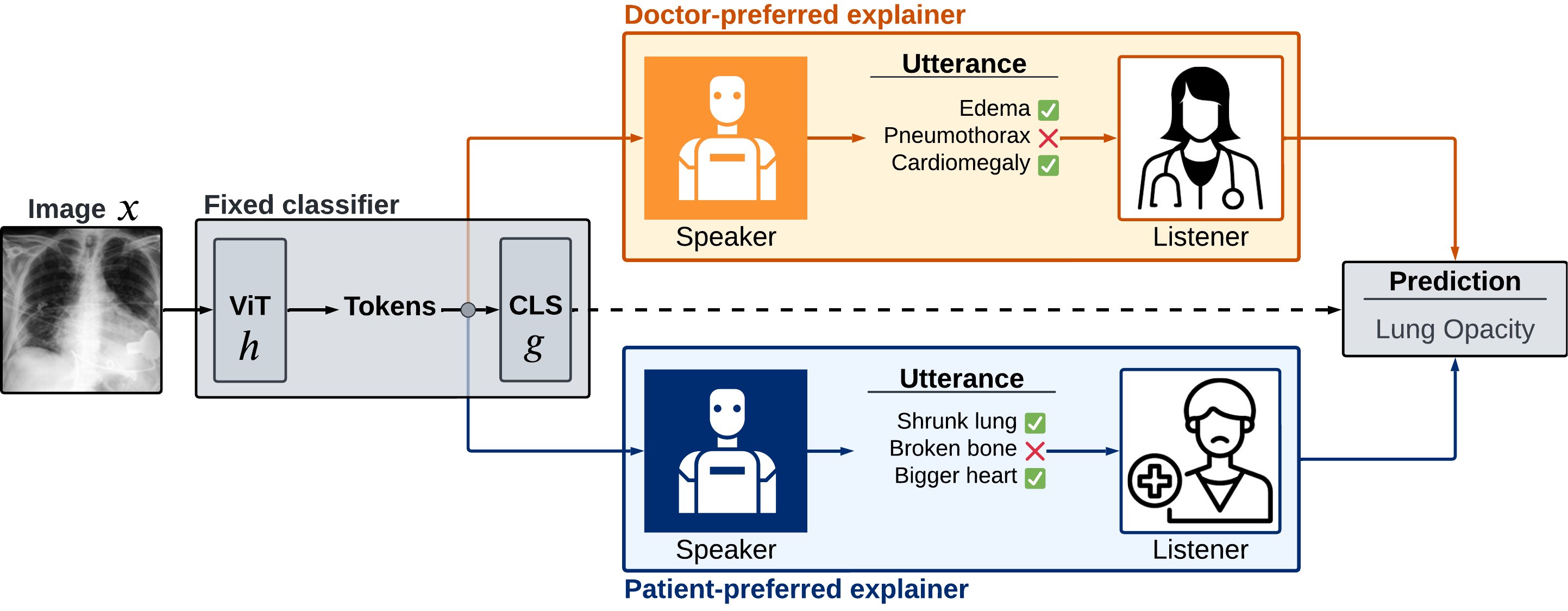}
    \caption{\label{fig:overview}Illustration of our listener-adaptive explanation framework: a speaker generates utterances to help a listener infer model predictions without access to the input image.}
\end{figure*}

\subsection*{\textbf{Summary of Contributions}}
\noindent We briefly summarize the main contributions of this work:
\begin{itemize}[leftmargin=*]
    \item We frame the problem of explaining the predictions of black-box models as that of finding utterances that maximize the utility of a listener, thus adapting such explanations to different contexts and stakeholders through pragmatic reasoning.
    \item We introduce an iterative procedure based on DPO \cite{rafailov2023direct} to generate listener-adaptive explanations (as illustrated in \cref{fig:overview}). Our approach does not require backpropagation through an explicit listener model, which is important for real-world scenarios where preference data can only be collected from human feedback.
    \item We study the effectiveness of our approach on image classification across three datasets. In particular, we simulate listeners with different preferences over explanations in terms of the semantic features of an input image or technical vocabulary. We verify that our method generates explanations that align with the listeners' preferences.
    \item We validate the importance of pragmatic explanations with a user study. We show that explanations generated with speakers that align well with the preferences of simulated listeners outperform existing explanation methods, improving the ability of the participants to infer the prediction of the base classifier.
\end{itemize}

Our work contributes to the development of machine learning explanation methods that can adapt to the unknown preferences of different users. We show that reinforcement learning techniques, such as DPO, present promising technical tools that remain underexplored in the interpretability research field. With a user study, we validate the effectiveness of our proposed method and stress the importance of incorporating pragmatic reasoning to compose concept-based explanations that are effective at communication.

\subsection*{Related Works}
Here, we summarize relevant works at the intersection of explanations for machine learning models and reasoning. Pragmatic reasoning and theory of mind (ToM) \cite{premack1978does,baker2009action,jara2016naive} are long-standing topics of research in computational linguistics and cognitive science, especially in referring expression generation (REG) \cite{krahmer2012computational,schlotterbeck2023incremental} and image captioning \cite{cohn2018pragmatically,ou2023pragmatic}. From this perspective, recent works have studied audience-aware language adaptation in interactive settings \cite{takmaz2023speaking,hawkins2019continual,greco2023she}, and reviewed their implications for human-machine interaction \cite{brandizzi2023toward}. For machine learning explainability, prior works have deployed similar ideas, although not explicitly mentioning connections with pragmatics. For example, \citet{treviso2020explanation} also modeled machine learning interpretability as a cooperative game between a speaker and a listener model. Similarly---for textual inputs---\citet{lei2016rationalizing} proposed to \emph{rationalize} the predictions of black-box models by jointly training an extractor and a predictor network. In this work, we formalize these ideas with principles of pragmatic reasoning.

Our approach uses DPO to generate explanations that adapt to the preferences of different listeners. While DPO has shown strong performance across various language modeling \cite{liu2025survey} and reasoning \cite{pang2024iterative,tu2025enhancing} tasks, to our knowledge, it has not been applied to machine learning interpretability prior to this work. Of note, \citet{wu2024discret} and \citet{wang2018reinforcement} use other reinforcement learning techniques different from DPO in the context of interpretable individual treatment effect estimation and recommendations systems, respectively.

Finally, we note that our approach is related to the work of \citet{peng2024pragmatic}, who study how to learn pragmatic feature preferences from human data. In this work, we do not concern ourselves with the reward learning problem and focus on adapting to fixed preferences instead. We also note the work of \citet{pu2024amortizing}, who leverage user rankings to amortize exact RSA inference.

\section{Background}
We begin by briefly summarizing the necessary notation and background on the rational speech act and direct preference optimization. Herein, we will denote $\U$ and $\W$ as the spaces of \emph{utterances} and \emph{world states}, respectively. Furthermore, we will treat speakers $S: \W \to \U$ and listeners $L: \U \to \W$ as functions that map world states to utterances and vice versa. In the context of this work, utterances correspond to explanations of the predictions of machine learning models, while world states represent the predictions being explained.

\subsection*{Rational speech act (RSA)\cite{frank2012predicting,goodman2016pragmatic,andreas2016reasoning,scontras2021practical}}
\noindent The RSA assumes that listeners and speakers are rational---they act according to their internal models of the world to achieve goals. Although this is not always true in practice, the RSA has proven pivotal in modeling the behavior of agents in contextual settings. In particular, it formalizes pragmatic reasoning as recursive probabilistic inference. Intuitively, the RSA generates a sequence of listeners and speakers of increasing \emph{order}, \ie~$L_0,S_1,L_1,S_2,L_2,\dots$ with respective probability distributions $P_{L_j}(w \mid u)$ and $P_{S_j}(u \mid w)$, $\forall j \in \N_0$.

Speakers and listeners are recursive agents because they depend on each other. In particular, the level-0 listener is defined as
\begin{equation}
    P_{L_0}(w \mid u) \propto \delta(u,w) P_{L_0}(w),
\end{equation}
where $\delta(u,w) = \1\{\text{utterance}~u~\text{is true in state}~w\}$. Subsequently, at order $j > 0$,
\begin{gather}
    P_{S_j}(u \mid w) \propto P_{L_{j-1}}(w \mid u) P_{S_j}(u),\\
    P_{L_j}(w \mid u) \propto P_{S_{j}}(u \mid w) P_{L_j}(w).
\end{gather}
Put into words, the level-0 (or literal) listener $L_0$ induces an initial distribution over states $P_{L_0}(w \mid u)$ that is proportional to the product of the literal meaning of an utterance $u$ (\ie, $\delta(u,w)$) and a prior $P_{L_0}(w)$. That is, given an utterance, the posterior distribution of the listener will strictly include the world states where the utterance is true, and those states only. Then, the distribution of utterances of the level-1 (or pragmatic) speaker $S_1$ integrates information from the literal listener and $P_{S_1}(u)$, a prior over utterances (\eg, conciseness, specific topics, or other preferences). This speaker is called pragmatic because its distribution depends on the literal listener $L_0$, meaning it is aware of its communicative context. Finally, a pragmatic listener $L_1$ replaces the literal meaning of $u$ with the likelihood of the pragmatic speaker $S_1$ (\ie, $\delta(u,w)$ vs. $P_{S_1}(u \mid w)$). As before, this listener is pragmatic because it has an internal model of the speaker. Naturally, one can iterate this pattern to induce higher-order agents $S_2,L_2,S_3,L_3$, etc. Briefly, we note that the prior distributions $P_{S_j}(u)$ and $P_{L_j}(w)$, $j \in \N_0$ may change, reflecting the fact that, over time, both listeners and speakers may update their beliefs.

\subsection*{Direct preference optimization (DPO) \cite{rafailov2023direct,rosset2024direct}}
Reinforcement learning from human feedback (RLHF) \cite{askell2021general,ouyang2022training,kaufmann2023survey} is a prominent approach for the alignment of large language models (LLMs) with human values that go beyond a model's task definition \cite{russell2019human,gabriel2020artificial,hadfield2016cooperative,bubeck2023sparks,shen2023large}. Instead of directly maximizing a reward with online reinforcement learning \cite{schulman2017proximal,schulman2015trust}, DPO reformulates the RLHF objective as an offline optimization problem without an explicit reward model---which would otherwise need estimation from data. Formally, denote $q = (w,u^+,u^-)$ a preference pair for state $w$, such that $u^+$ is the chosen utterance and $u^-$ is the rejected one. The classical Bradley-Terry (BT) model \cite{zermelo1929berechnung,bradley1952rank} stipulates that
\begin{equation}
    \P[u^+ \succ u^- \mid w] = \sigma(r^*(u^+,w) - r^*(u^-,w)),
\end{equation}
where $r^*$ is the optimal reward model, and $\sigma$ is the sigmoid function. That is, the probability of $u^+$ being preferred to $u^-$ is proportional to their reward difference. \citet{rafailov2023direct} show that even though $r^*$ may be unknown (\eg, because it encodes human preferences), it can be rewritten in terms of the optimal RLHF speaker $S^*$. In particular, it holds that $r^*(u,w) \propto \beta \log(P_{S^*}(u \mid w)/P_{S_{\rref}}(u \mid w))$, where $S_{\rref}$ is a reference model (usually a fine-tuned LLM), and $\beta$ is the strength of the regularizer $\KL(S^* || S_{\rref})$. Then, the DPO objective is the maximum likelihood estimate of $r^*$ under the BT model. Denote $\D$ the distribution of $q = (w,u^+,w^-)$, then 
\begin{equation}
    \label{eq:dpo}
    \L_{\DPO}(S,S_{\rref}) = - \E_{q \sim \D}~\left[\log\sigma\left(\Delta(q) \right)\right],
\end{equation}
where
\begin{equation}
    \Delta(q) = \beta \left(\log\frac{P_S(u^+ \mid w)}{P_{S_{\rref}}(u^+ \mid w)} - \log\frac{P_S(u^- \mid w)}{P_{S_{\rref}}(u^- \mid w)}\right).
\end{equation}
We stress that optimizing this loss only requires access to a dataset of preferences $\{(w_i, u^+_i, u^-_i)\}_{i=1}^n$ and it does not explicitly model $r^*$, which may be difficult to do in practice.

\smallskip
\noindent With this background, we now introduce the main contributions of our work.

\section{Pragmatic Explanations}
The goal of post-hoc explainability is to elucidate the mechanisms of complex black-box predictors such as modern neural networks. For example, existing research has focused on developing feature attribution methods using different notions of importance, such as gradients \cite{selvaraju2016grad,wang2024gradient,kolek2022cartoon}, game- and information-theory \cite{lundberg2017unified,kolek2020rate,kolek2022cartoon,teneggi2022fast,covert2021explaining}, or conditional independence \cite{teneggi2022shap,burns2020interpreting,teneggi2024testing,shi2024hypothesis,bharti2024sufficient}. Furthermore, mechanistic interpretability \cite{cunningham2023sparse,nanda2023progress,conmy2023towards,bereska2024mechanistic,dunefsky2024transcoders} has emerged as a popular approach to dissecting the internal mechanisms of frontier language models to identify circuits that encode various capabilities. All these methods, however, generally overlook the \emph{context} of explanation---fundamental for effective communication with the users of a model. Following the RSA, we argue that an explanation is communicative if a listener can use it to infer the predictions made by a model.

More formally, let $f: \X \to \Y$ with $\Y = [k] \coloneqq \{1, \dots, k\}$ be a fixed predictor that maps an input $x \in \X$ to one of $k$ classes. We assume $f$ can be seen as the composition of a feature encoder $h: \X \to \R^d$ with a downstream classifier $g: \R^d \to \Y$, \ie~$f(x) = g(h(x))$, and that we have access to the embeddings $h(x)$. We define $S: \R^d \to \U$ as a speaker model (\ie, the explainer) that takes embeddings as input and generates utterances in $\U$. Following prior work on concept-based explanations \cite{koh2020concept,kim2018interpretability,yuksekgonul2022post}, we consider utterances (\ie, explanations) composed of at most $l$ claims from a vocabulary $\C = \{c_1,\dots,c_m\}$, where each claim can be labeled \verb|+1| or \verb|-1| depending on whether it is true or false for input $x$, \ie~$\U \subseteq (\C \times \{-1,+1\})^{\leq l}$. For example, in chest X-ray classification, $f$ may predict whether a scan shows signs of lung opacity, and $\C$ is a vocabulary of medical findings such as \emph{edema}, \emph{consolidation}, or \emph{pneumothorax}. Then, a possible utterance explaining the prediction \emph{``signs of lung opacity''} could be $u = [(\verb|edema,+1|), (\verb|pneumothorax,-1|)]$. 

We remark that this choice is not unique, and other types of utterances exist, such as freeform or counterfactual explanations \cite{guidotti2024counterfactual}. While limited, this will be a useful choice motivated by the fact that: $(i)$ concept-based explanations allow for comparison with existing interpretability methods, such as V-IP \cite{chattopadhyay2023variational}; $(ii)$ they can be grounded in existing datasets with human annotations of semantic features \cite{irvin2019chexpert,wah2011caltech,daneshjou2022skincon}; and $(iii)$ they can be presented intuitively as a list of facts---which will be important for the design of our user study. In contrast, freeform explanations would require claim decomposition and verification \cite{wanner2024dndscore}, an additional layer of complexity outside the scope of the current submission, and counterfactual explanations may put a high cognitive load on the participants of our user study. Specifically, it would have been infeasible for participants to meaningfully understand counterfactual claims between hundreds of possible classes.

To connect machine learning explanations with pragmatic reasoning, let $L: \U \to \Y$ denote a listener that receives an utterance and responds with one of the possible classes. We remark that this listener need not be a neural network, or, more generally, a differentiable model, and that it can represent a human. In the context of this work, we will simulate listeners with bidirectional transformers, but this is only a practical choice to study the effectiveness of our training procedure, given the limited availability of human preference data (and we consider collecting data from human feedback as important future work). Following the RSA, we propose that the distribution of utterances induced by a pragmatic explainer should balance \emph{consistency} with the input and \emph{utility} for a listener. That is, for an input $x$ and prediction $\y = f(x)$, the distribution of utterances induced by a pragmatic explainer should should satisfy
\begin{equation}
    \label{eq:speaker_likelihood}
    \log P_S(u \mid h(x)) \propto \underbrace{\log P(u \mid x)}_{\text{consistency}} + \underbrace{\log P_L(\y \mid u)}_{\text{utility}},
\end{equation}
where the first term represents the consistency of the utterance with the input $x$, and the second is the utility of the utterance for the listener. We briefly note that consistency differs from the notion of \emph{faithfulness} in the broader interpretability literature. The former simply refers to the agreement between the contents of an input and the utterance of a speaker, whereas the latter refers to explanations being representative of the inner mechanism of a black-box predictor. We will expand on the implications of this difference later in the manuscript. 

\cref{eq:speaker_likelihood} considers a fixed listener who is able to infer $\y$ rationally, but it is common for listeners to only have partial knowledge about the task at hand in many real-world scenarios. Going back to our example of doctor and patient listeners, the former knows the underlying pathophysiology of a disease, but the latter might not. In these cases, listeners will naturally learn over time by interacting with the speaker \cite{takmaz2023speaking}. In order for our approach to account for the adaptation of both speakers and listeners, we frame our learning objective as a joint optimization problem, where the goal is to maximize communicative reward. That is
\begin{equation}
    \label{eq:objective}
    (S^*, L^*) = \argmax_{S,L} \underset{\substack{(x,\y) \sim \D_f \\ u \sim P_S(\cdot \mid h(x))}}{\E}[R_{\alpha}(x,\y,u)],
\end{equation}
where
\begin{equation}
    \label{eq:reward}
    R_{\alpha}(x,\y,u) = \log P(u \mid x) + \alpha \log P_L(\y \mid u),
\end{equation}
$\D_f$ is the joint distribution of $(x,\y = f(x))$, and $\alpha \geq 0$ is a hyperparameter that controls the tradeoff between consistency and utility (when $\alpha = 0$, the listener is ignored and the speaker is literal, maximizing utterance consistency only).

We remark that our setup is related to existing works on building \emph{inherently-interpretable} predictors \cite{chattopadhyay2023variational,chattopadhyay2024bootstrapping} and on \emph{rationalizing} the predictions of black-box classifiers \cite{treviso2020explanation,lei2016rationalizing,liu2024mmi}. In particular, \citet{treviso2020explanation} also consider a cooperative game between an explainer and a layperson listener, modeled with neural networks that are trained jointly. Similarly, \citet{chattopadhyay2023variational} show how to train interpretable predictors by maximizing the mutual information between the ground-truth label $y$ and the next claim $c \in \C$ to include in an utterance $u$. Therefore, their method, V-IP, can be seen as a solution to \cref{eq:objective} when the listener model is known and differentiable. Unlike existing works, we develop an algorithm that does not need differentiation through a listener model, which is important for scenarios where preference data is easier to collect than estimating an explicit reward model. Finally, we note that the utility term $\log P_L(\y \mid u)$ quantifies the \emph{sufficiency} of the utterance for the listener to infer the predicted label, and previous works have studied this notion in the context of machine learning explainability. Specifically, \citet{kolek2020rate} and \citet{bharti2024sufficient} study sufficiency (and necessity) from an information-theoretic and statistical perspective, respectively, where the listener is the classifier itself, and the goal is to find a subset of the input features $s \subseteq [d]$ such that $f(x) \approx f(x_{s})$.

\begin{algorithm}[t]
    \caption{\label{algo:procedure}PragmatiXs: Training pragmatic explainers}
    \raggedright\textbf{Inputs:} Dataset $D = \{(x_i,\y_i)\}_{i=1}^n$, fixed encoder $h: \X \to \R^d$, maximum utterance length $l$, pragmatics strength $\alpha \geq 0$, true negative weight $\gamma \in [0,1]$, DPO regularization strength $\beta \geq 0$, number of utterances $n_{\expl}$, $n_{\pref}$, and number of iterations $T$.
    \begin{algorithmic}[1]
        \STATE $S\at{0}, L\at{0} \gets \text{random initialization}$
        \FOR{$t = 1, \dots, T$}
            \STATE $D_{\pref}\at{t} \gets \textsc{Preferences}(D,S\at{t-1},L\at{t-1},\alpha,n_{\pref})$
            \STATE $S\at{t} \gets \textsc{DPO}(D_{\pref}\at{t},S\at{t-1},\beta)$
            \STATE $D_{\expl}\at{t} \gets \textsc{Explanations}(D,S\at{t},n_{\expl})$
            \STATE $L\at{t} \gets \textsc{NLL}(D_{\expl}\at{t},L\at{t-1})$
        \ENDFOR
        \RETURN $S\at{T},L\at{T}$
    \end{algorithmic}
\end{algorithm}

\subsection{The Training Procedure}
In this section, we present our training procedure. We remark that, for the purpose of the algorithm and our experiments, we simulate listeners with neural networks for two reasons: first, human preference data remains scarce for machine learning explainability, and, second, they allow us to study the effectiveness of our approach with listeners that evolve over time, important for real-world applications.

We follow an alternating optimization approach to approximate a solution to \cref{eq:objective} (summarized in \cref{algo:procedure}). In particular, the speaker and listener models are initialized at random and, at each iteration $t = 1, \dots, T$, they are updated sequentially. Denote $D = \{(x_i,\hat{y}_i)\}_{i=1}^n$ a dataset of inputs with their respective predictions by a base classifier $f$ we want to explain. Then:
\begin{enumerate}[leftmargin=*]
    \item Given a fixed listener $L\at{t-1}$, the new speaker $S\at{t}$ can be found via DPO on $D_{\pref}\at{t} = \{(x_i, \{(u^+_j, u^-_j)\}_{j=1}^{n_{\pref}})\}$, where $n_{\pref} = b(b-1)/2$ is the number of pairwise preferences constructed by ranking $b$ utterances from $S\at{t-1}$ according to \cref{eq:speaker_likelihood}. Note that, at each step, we use $S\at{t-1}$ as the reference model in the DPO objective to regularize the update and avoid drastic changes.
    \item Given a fixed speaker $S\at{t}$, its optimal listener $L\at{t}$ can be estimated from $L\at{t-1}$ via cross-entropy minimization over a dataset of explanations $D_{\expl}\at{t} = \{(\y_i, (u_{i,1}, \dots, u_{i,n_{\expl}}))\}_{i=1}^n$, where $n_{\expl}$ is the number of utterances for each input $x_i$, i.e. $u_{i,j} \sim P_{S\at{t}}(\cdot \mid h(x_i))$.
\end{enumerate}
Note that using DPO to update the speaker avoids backpropagation through the listener model. This is important for real-world scenarios with human subjects, whose reward may be difficult to estimate explicitly, but preference data can be collected. We will verify that using simulated listeners improves the classification performance of participants of a user study in \cref{sec:human_evaluation}.

\subsection*{Grounding explanations}
\noindent Recall that the consistency term in the reward (\cref{eq:reward}) should ground utterances in a knowledge base, such as image captions, expert annotations, or the likelihood of a pretrained vision-language model. Here, we define
\begin{equation}
    P(u \mid x) = \frac{\exp(\cons(u,x))}{\sum_{u'} \exp(\cons(u',x))},
\end{equation}
where $\cons(u,x) \in [0,1]$ is an unnormalized score. This way, the pairwise ranks of candidate utterances do not depend on the normalizing constant $Z(x) = \sum_{u'} \exp(\cons(u',x))$ but on $\cons(u,x) + \alpha P_L(\y \mid u)$ only. 

Given the binary nature of concept-based claims---and as done in previous works \cite{koh2020concept,kim2018interpretability,chattopadhyay2023variational,chattopadhyay2024bootstrapping}---we assume access to \emph{semantics} $z: \X \to \{-1,0,+1\}^m$ in the form of ground-truth annotations or predictions from a vision-language model (e.g., CLIP \cite{radford2021learning}, Concept-QA \cite{chattopadhyay2024bootstrapping}). These semantics $z(x) \in \{-1,0,+1\}^m$ indicate which claims are true for $x$ (i.e., $z(x)_j = +1$), which are false (i.e., $z(x)_j = -1$), and which are unknown (i.e, $z(x)_j = 0$). Then, we define $\cons$ as the weighted average of the number of true positive claims with the number of true negative claims, i.e. $\TP = \lvert\{(c_j, \z_j) \in u: \z_j = 1 \wedge z(x)_j = 1\}\rvert$, $\TN = \lvert\{(c_j, \z_j) \in u: \z_j = -1 \wedge z(x)_j = -1\}\rvert$, and
\begin{equation}
    \label{eq:fidelity}
    \cons(u,x) = \frac{\TP + \gamma\TN}{\lvert u \rvert},~\gamma \in [0,1].
\end{equation}
As $\gamma \to 0$, utterances with negative claims are downvoted, and the opposite is true as $\gamma \to 1$. Varying $\gamma$ accounts for the fact that, as the vocabulary $\C$ grows, it is easier for a claim $c \in \C$ to be false, making the ground-truth semantics imbalanced. Finally, we remark that the ground-truth semantics are used to compute consistency and rank utterances only. The speaker does not receive direct supervision of which claims in an utterance are correct or not, and instead, it only receives weak supervision through preference pairs that encode consistency information. This is important for real-world applications with human subjects beyond concept-based explanations, where consistency may be hard to estimate directly from data.

\subsection*{Simulating listener preferences}
\noindent In our experiments, we simulate listeners with different preferences to verify that our algorithm can successfully align speakers and their explanations. There are several ways to achieve this by inducing preferences over particular topics, categories, or styles of presentation (\eg, technical nomenclature vs. common parlance). Here, we use temperature scaling \cite{guo2017calibration} with respect to a prior distribution over \emph{groups} of claims. That is, we assume access to a family of groups $\G = \{g_j: \C \to \{0,1\}\}$ such that $g_j(c) = 1$ if $c$ belongs to the $j\th$ group (and groups may intersect). Then, a listener $L_{\pi}$ has a prior distribution over groups $\pi \in \Delta^{\lvert \G \rvert}$ and
\begin{equation}
    \label{eq:temperature_scaling}
    P_{L_{\pi}}(\y \mid u) = \texttt{softmax}\left(\frac{\xi(u)}{\tau \cdot \KL(g(u) || \pi) + 1}\right),
\end{equation}
where $\xi: \U \to \R^{k}$ are the unnormalized outputs (the logits) of the neural network simulating the listener, $g: \U \to \Delta^{\lvert \G \rvert}$ is the distribution of groups in utterance $u$, and $\tau \geq 0$ is a temperature parameter that strengthens the effect of the prior. Intuitively, the larger the distance between the observed and prior distributions, the more uniform $P_{L_{\pi}}(\y \mid u)$ becomes, making it harder for the listener to predict $\y$ confidently. We note that we use the reverse KL in order to penalize utterances with claims belonging to groups that have zero mass in the prior (i.e., $\pi_j = 0$ and $g(u)_j > 0$), which would not have an effect on the forward KL.

\section{Experiments}
We evaluate and compare our method on three image classification datasets: CUB \cite{wah2011caltech}, ImageNet \cite{deng2009imagenet,russakovsky2015imagenet}, and CheXpert \cite{irvin2019chexpert}. For each dataset, we use the zero-shot classifier of a pretrained vision-language model as the base predictor. In particular, we use OpenClip:ViT-L-14 \cite{cherti2023reproducible,radford2021learning,schuhmann2022laion} for CUB and ImageNet, and BiomedVLP \cite{bannur2023learning} for CheXpert. We briefly summarize the classification problem for each dataset and the semantic concepts:

\paragraph{CUB (200 classes and 312 concepts)}
The CUB dataset comprises 11,788 images of 200 different bird species, where each image is annotated with 312 human-interpretable concepts that describe the appearance of the bird. We use all classes and concepts from the original dataset, and we classify images with the zero-shot classifier obtained by encoding \texttt{"A photo of a <BIRD\_NAME>"} with the text encoder of the vision-language model. We use 9,430 images for training and 2,358 for testing.

\paragraph{ImageNet (300 classes and 400 concepts)}
ImageNet is a landmark dataset of approximately 14 million images of common objects sourced from the internet. We downsample the ImageNet-1K subset by randomly selecting 300 classes, and, for each class, 100 images for training and 50 for testing (30,000 and 15,000 images in total). Since ImageNet does not provide concept annotations, we use Concept-QA \cite{chattopadhyay2023variational} to annotate a vocabulary of 4,523 claims, of which we keep the 400 most prevalent in the training set. Similarly to above, the zero-short classifier is the encoding of \texttt{"A photo of a <CLASS\_NAME>"}.

\paragraph{CheXpert (2 classes and 12 concepts)}
CheXpert is a large-scale dataset of 224,316 chest X-rays with their radiology reports. Each report is automatically labeled for the presence of 14 clinical findings extracted from the clinical report. We predict whether a scan contains any signs of lung opacity with the zero-shot classifiers \texttt{"No signs of lung opacity"} and \texttt{"Findings suggesting lung opacity"}. We train the models on the first 10,000 patients in the dataset, for a total of approximately 40,000 images. We use the original set of 234 scans for testing, and the VisualCheXbert labels \cite{jain2021visualchexbert} as ground-truth semantics. We remark that semantics for the test set are verified by trained radiologists. Since lung opacity is one of the clinical findings, we exclude it from the list of semantic concepts. We also remove the concept "no findings" since it overlaps with the classification problem, for a total of 12 concepts.

\smallskip
We implement speakers as multimodal encoder-decoder models following the CoCa architecture \cite{yu2022coca} and we simulate listeners with bidirectional transformers \cite{devlin2019bert} (see Appendix~\ref{supp:models} for details). We train all models with AdamW \cite{loshchilov2017decoupled} with a learning rate of 0.0001, weight decay of 0.01, gradient norm clipping at 1.0, and cosine annealing \cite{loshchilov2016sgdr} after each iteration of \cref{algo:procedure}. 

The hyperparameters of our procedure balance several tradeoffs, such as the ratio of positive to negative claims in an utterance, the weight of the prediction of the listener in the reward, and computational constraints. In our experiments, we use $\alpha = 0.2$ when training pragmatic speakers, true negative weight $\gamma = 0.4$, DPO regularization strength $\beta = 0.6$, number of utterances $n_{\expl} = 8$, and number of candidate utterances $b = 4$ (i.e., $n_{\pref} = b(b-1)/2 = 6$). We refer interested readers to Appendix~\ref{supp:hyperparameters} for details on hyperparameters and their effect on the training procedure.

Herein, we will refer to listener models trained with literal speakers (\ie, $\alpha = 0$ in \cref{eq:reward}) as \emph{literal listeners}, and to those trained with pragmatic speakers (\ie, $\alpha > 0$ in \cref{eq:reward}) as \emph{pragmatic listeners}. We now present and discuss the main findings of our experiments.

\begin{table}[t]
\caption{\label{table:accuracy}Summary of results across all datasets. We include both listener prediction accuracy and explanation accuracy in parentheses. Literal and pragmatic speakers are trained with $\gamma = 0.4$ and $\alpha = 0.0,0.2$, respectively. (*) V-IP achieves an explanation accuracy of 100\% on ImageNet because we use the predicted labels of Concept-QA as ground-truth, since the dataset does not include annotations of semantic concepts. We could not compare with V-IP on CheXpert because there is no pretrained concept predictor. We did not train pragmatic speakers with warm-up on CheXpert because the vocabulary contains 12 concepts only.}
\centering
\small
\resizebox{\linewidth}{!}{%
\begin{tabular}{lcccccccc}
\toprule
                            &                       &                               &                           &                           & \multicolumn{4}{c}{prediction accuracy \acc{explanation accuracy}}\\
                                                                                                                                            \cmidrule(r){6-9}                                                                              
Dataset                     & classes               & concepts                      & \mline{base\\accuracy}    & \mline{utterance\\length} & V-IP                              & literal               & pragmatic                         & \mline{pragmatic\\\scriptsize(warm-up)} \\
\midrule
\multirow{2}{*}{CUB}        & \multirow{2}{*}{200}  & \multirow{2}{*}{312}          & \multirow{2}{*}{75.70\%}  & 6                         & 26.17\%~\acc{72.29\%}             & 13.91\%~\acc{79.05\%} & 52.80\%~\acc{76.38\%}             & \textbf{59.25\%}~\acc{76.53\%} \\
                            &                       &                               &                           & 30                        & \textbf{69.80\%}~\acc{70.82\%}    & 42.79\%~\acc{77.73\%} & 61.37\%~\acc{76.21\%}             & 64.55\%~\acc{75.25\%} \\ 
                            \arrayrulecolor{gray}\midrule\arrayrulecolor{black}
\multirow{2}{*}{ImageNet}   & \multirow{2}{*}{300}  & \multirow{2}{*}{400}          & \multirow{2}{*}{82.11\%}  & 12                        & 51.05\%~\acc{100\%$^*$}           & 33.72\%~\acc{88.70\%} & 67.10\%~\acc{86.94\%}             & \textbf{70.55\%}~\acc{87.81\%}\\
                            &                       &                               &                           & 30                        & 65.51\%~\acc{100\%$^*$}           & 51.25\%~\acc{88.10\%} & 67.80\%~\acc{86.04\%}             & \textbf{73.91\%}~\acc{88.24\%}\\
                            \arrayrulecolor{gray}\midrule\arrayrulecolor{black}
CheXpert                    & 2                     & 12                            & 78.21\%                   & 4                         & -                                 & 85.90\%~\acc{81.20\%} & \textbf{99.15\%}~\acc{75.93\%}    & -\\
\bottomrule
\end{tabular}}
\end{table}

\subsection{Pragmatic explanations improve listener accuracy}
Here, we evaluate whether jointly training speakers and listeners improves prediction accuracy, meaning that pragmatic explanations are more communicative than literal ones. We remark that literal listeners are updated following \cref{algo:procedure} even though they do not play a role in ranking utterances (\ie, $\alpha = 0$). This is important to guarantee a fair comparison with pragmatic listeners---precisely quantifying the gap in communicative utility between pragmatic and literal explanations. We enforced literal speakers to generate utterances with fixed length to prevent the collapse of the explanations, as short utterances trivially maximize consistency. To compare with V-IP \cite{chattopadhyay2023variational}, we followed the original implementation and trained interpretable models that infer the predictions of the base classifiers instead of the ground-truth labels. We could not compare with V-IP on CheXpert because a pretrained concept classifier is not available. \cref{table:accuracy} summarizes results across all datasets, including the accuracy of the base classifier, the accuracy of the predicted labels of the claims in the utterances, and the prediction accuracy of the listeners.

\begin{figure*}[t]
    \centering
    \includegraphics[width=\linewidth]{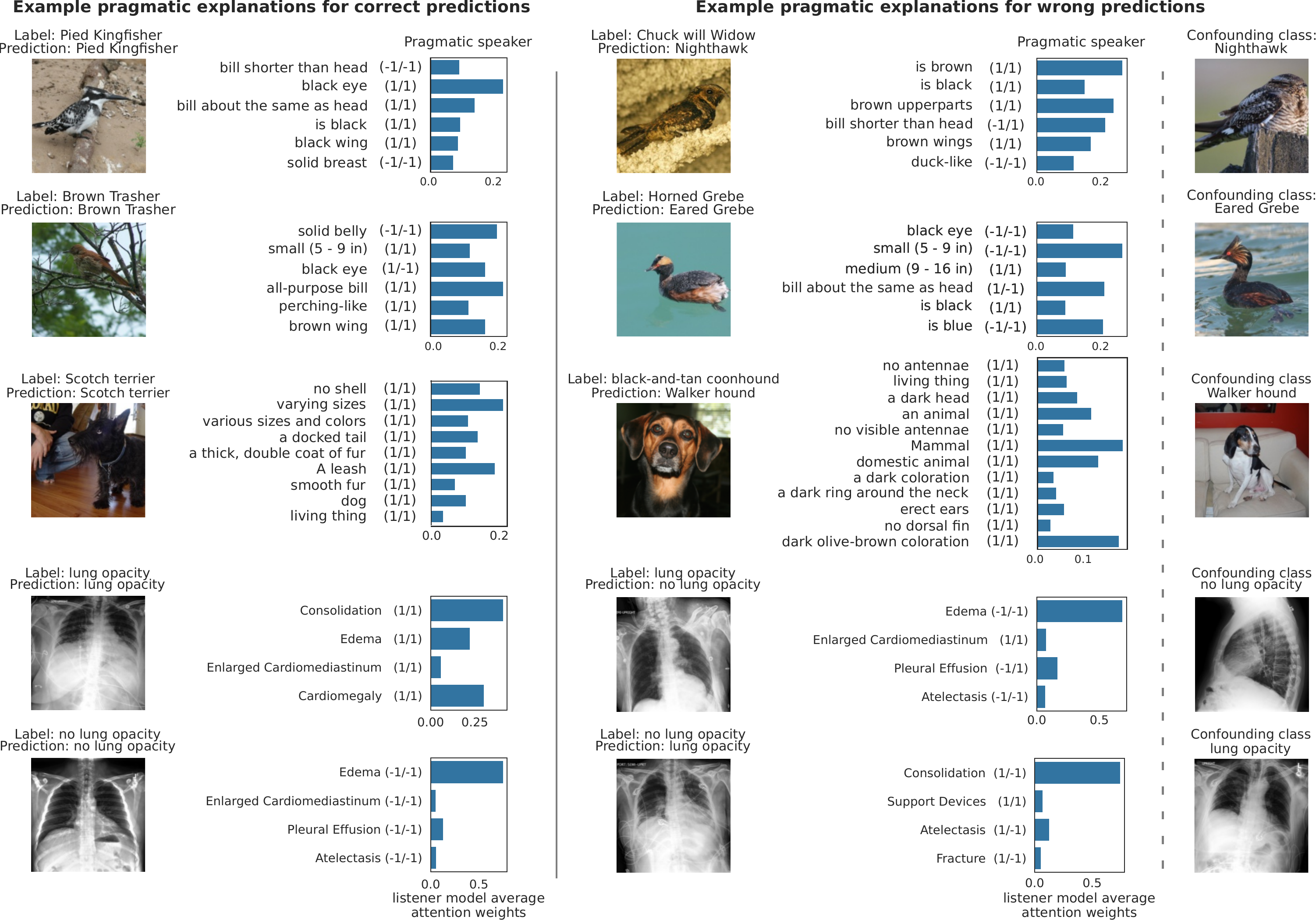}
    \caption{\label{fig:cub_examples}Example utterances generated with pragmatic speakers on all datasets. We include the input image, the utterance with the predicted and ground-truth labels in parentheses, and the average attention weights of the listener model across all layers. The first panel shows examples where the base classifier was correct, and the pragmatic listener inferred the predicted label. The second panel shows examples where the base classifier was wrong, and the pragmatic listener inferred the wrong predicted label. We include an example image of the confounding class to compare with.}
\end{figure*}

First, we can see that pragmatic explanations improve prediction accuracy compared to literal explanations. This trend supports the motivation of our work: considering pragmatic reasoning when developing machine learning explanation methods leads to better communication ability. We note that pragmatic explanations outperform V-IP in terms of prediction accuracy on the CUB dataset with utterances with 6 claims, and on ImageNet across all lengths studied in our experiments. However, V-IP performs better than our method with utterances longer than 6 claims on CUB (we include results for 30 claims in the table). This agrees with the intuition that V-IP is a solution to the optimization problem in \cref{eq:objective} when the listener model is known and differentiable. Our approach, instead, uses reinforcement learning techniques that rely on exploration of the utterance space---which grows exponentially in the number of claims. To address this limitation, we propose a variation of our training procedure with a random warm-up phase. Specifically, when building the preference dataset $D\at{t}_{\pref}$, we mix utterances generated with the speaker model $S\at{t-1}$ with random utterances. We anneal the mixing rate linearly from 1.0 to 0.0 over the first half of the iterations, simultaneously increasing the learning rate. This promotes diversity of utterances, preventing the speaker and listener models from converging too fast to a suboptimal solution. On both the CUB and ImageNet datasets, we found that pragmatic speakers with a random warm-up phase improved listener performance, especially with long utterances. Finally, we include results as a function of utterance length and true negative weight $\gamma$ on the CUB and CheXpert datasets in Appendix~\ref{supp:accuracy}, and we investigate the effects of increasing the number of candidate utterances in our sampling procedure (\ie, $n_{\pref}$ and $n_{\expl}$) in Appendix~\ref{supp:hyperparameters}.

\begin{figure}[t]
    \centering
    \includegraphics[width=\linewidth]{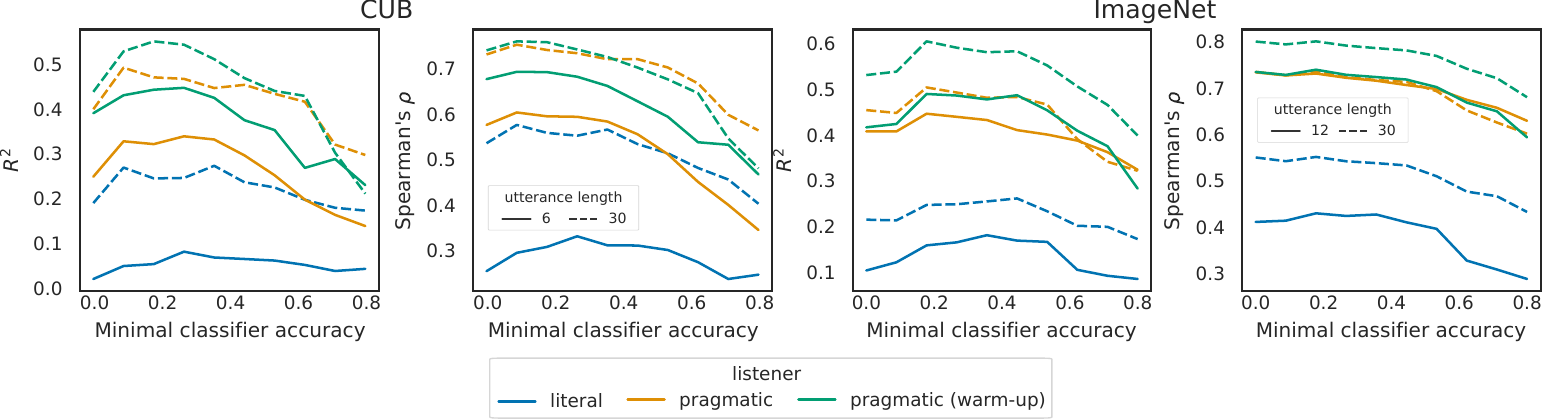}
    \caption{\label{fig:correlation}Correlation between the class-wise accuracy of the base classifier and of listener models. Results are shown as a function of utterance length and minimal base classifier accuracy.}
\end{figure}

Second, we compare the accuracy of explanations generated with literal and pragmatic speakers with those with V-IP. We stress that the purpose of this comparison is to investigate whether weak supervision via preference pairs can lead to comparable performance to strong supervision with concept annotations. We found that for the CUB dataset, explanations generated with our method are similar to those with V-IP (literal and pragmatic explanations are $\approx 6\%$ more accurate). We remark that the authors of \cite{chattopadhyay2023variational} preprocess the labels in the CUB dataset before training, whereas we do not, and that evaluation is affected by this difference. Furthermore, we evaluate the accuracy of the claims that are included in the utterances only, which are a small subset of the complete vocabulary of concepts in the dataset. For ImageNet, instead, recall that we use the predicted labels of Concept-QA as the ground-truth annotations, since the dataset does not include semantic concepts. Therefore, V-IP trivially achieves an accuracy of $100\%$. We found that both literal and pragmatic explanations achieve good classification performance ($\approx 87\%$). Importantly, we note that the accuracy of pragmatic explanations is only marginally lower than literal ones ($\approx 2\%$ on CUB and ImageNet, and $\approx 6\%$ on CheXpert), indicating an effective tradeoff between consistency and utility. Finally, the random warm-up phase slightly hurt explanation accuracy on CUB, while improving on ImageNet. 

Third, we remark that pragmatic explanations may convey wrong predictions, since communicative reward is defined with respect to predicted labels. That is, a listener may correctly infer the wrong predicted label. This is undesirable, as it may lead to deceptive explanations that instill a false sense of trust in the users of a black-box model. Indeed, prior work \cite{han2023ignorance} has shown that expert users have strong cognitive biases when considering post-hoc explanations of machine learning models. We remark that the focus of this work is on developing a technical framework for adaptive explanations, but these are important considerations for the safe deployment of our ideas in high-stakes scenarios. For example, \emph{learning with rejection} from human-in-the-loop machine learning \cite{cortes2016learning,jiang2018trust,charoenphakdee2021classification,hendrickx2024machine} could be used to limit explanations to confident predictions. Distribution-free uncertainty quantification techniques such as conformal prediction may be coupled with these to obtain statistical guarantees on the behavior of the model \cite{szabadvary2025classification,tayebati2025learning}. We consider this as follow-up work. \cref{fig:cub_examples} includes example pragmatic explanations on all datasets, for both correct and wrong predictions. We include the input image, the utterance generated with the speaker, and the average attention weights of the listener model. For wrong predictions, we also include an example image from the confounding class. Qualitatively, in the case of wrong predictions for the CUB and ImageNet datasets, the confounding classes are similar to the target, with several shared features. In these cases, pragmatic explanations with only a few concepts may not be able to disambiguate the two, or, alternatively, either the vocabulary of concepts or the pretrained embeddings may not be expressive enough. For the CheXpert dataset, instead, there are only two classes that do not share semantic attributes. Here, inspecting the attention weights of the listener model paints a different picture. We stress that the attention weights do not constitute well-grounded explanations, but they can provide insights into the internal representation of the listener model. Specifically, for the abnormal scan labeled as healthy, the attention weights concentrate on the \texttt{"no edema"} claim. We can see that this pattern matches the attention weights in the case of a truly healthy scan. This suggests that the listener model has learned to map the absence of edema to a negative prediction. Indeed, even though from a clinical perspective the absence of edema is not sufficient to rule out lung opacity on its own, over 97\% of negative scans in the training split of the dataset do not have edema. A similar situation arises with the healthy scan labeled as ``with lung opacity'': the speaker wrongly predicts that the scan contains findings of consolidation (\ie, an abnormal part of the lung that might be fluid or a mass), and the high attention weight on this claim suggests the listener model has learned to map the presence of consolidation to lung opacity (which is correct from a clinical perspective).

\begin{figure}[t]
    \centering
    \includegraphics[width=0.5\linewidth]{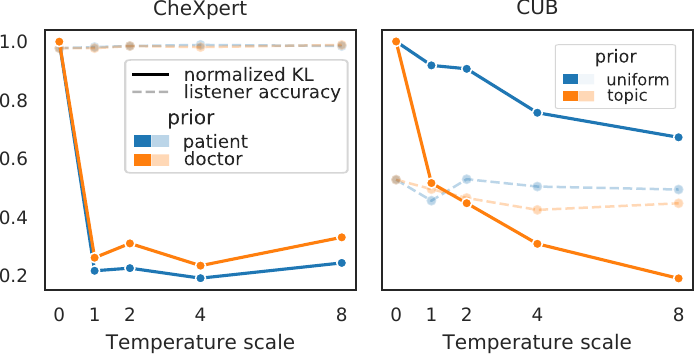}
    \caption{\label{fig:adaptation_kl}Utterance adaptation results on the CheXpert and CUB datasets. Solid lines report the normalized KL divergence between the empirical distribution of groups of claims in the utterances generated with a pragmatic speaker and the listeners' priors. Dashed lines report listener accuracy. Results are shown as a function of the temperature scale $\tau$.}
\end{figure}

\begin{figure*}[t]
    \centering
    \includegraphics[width=\linewidth]{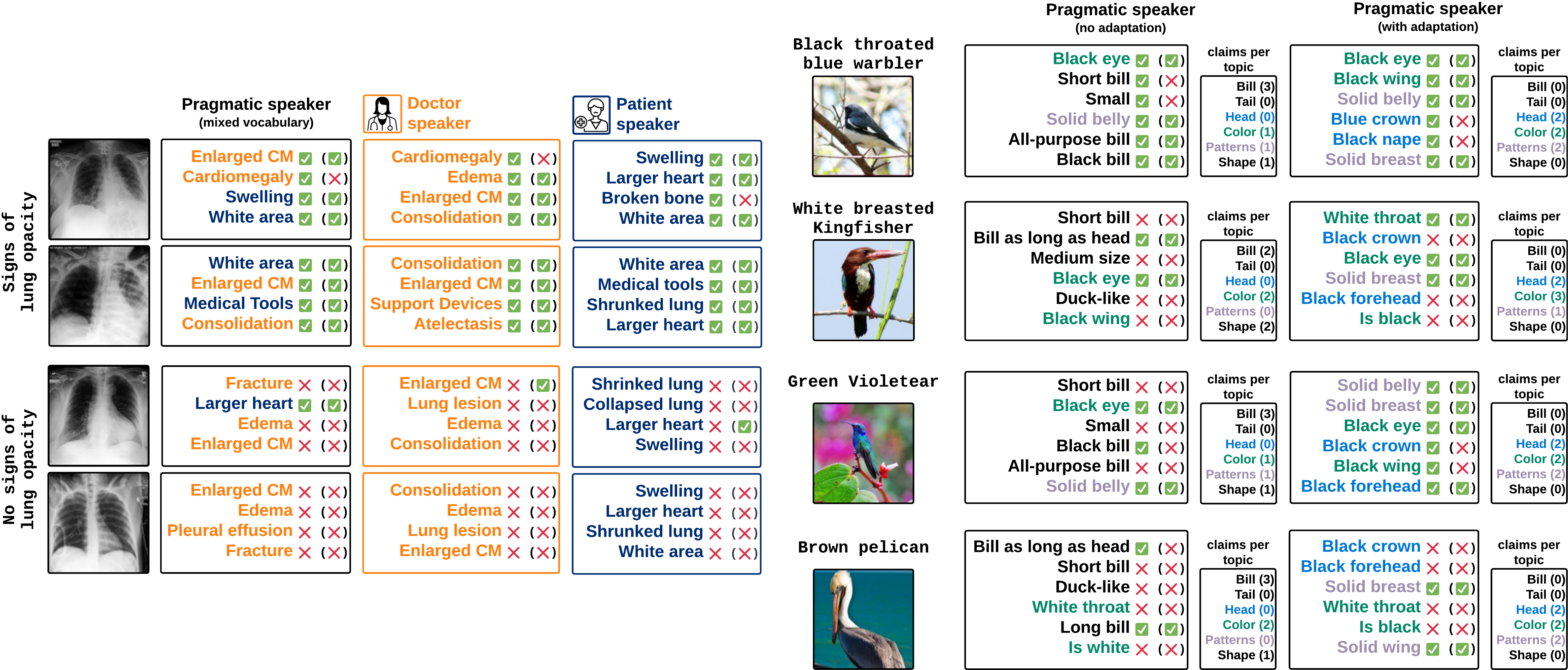}
    \caption{\label{fig:adaptation_example}Example utterance generated with pragmatic speakers with no adaptation and adaptation on the CheXpert (top row) and CUB (bottom row) datasets. We include examples where the base classifier and all listeners are correct in their respective predictions. For CheXpert, we include utterances generated with a pragmatic speaker with no adaptation on the augmented vocabulary of medical and layman's terms, and with pragmatic speakers adapted to doctor and patient listeners, respectively. For CUB, we include utterances generated with a pragmatic speaker with no adaptation, and with a pragmatic speaker adapted to a listener with a topic prior that excludes claims about bill, tail, and shape features.}
    \vspace{-10pt}
\end{figure*}

\subsection{Pragmatic listeners correlate with the base classifier}
It is natural to ask to what extent pragmatic explanations reflect the underlying predictor. First, beyond the consistency term in \cref{eq:speaker_likelihood}---which grounds utterances in a knowledge base---speakers are free to generate any utterance. Consistency is complementary to the notion of faithfulness in the broader interpretability literature, which refers to explanations being representative of a model's inner mechanism \cite{vilone2021notions,montavon2018methods}. This distinction is important to avoid harmful consequences in high-stakes scenarios, and prior works have studied the roles of \emph{plausibility vs. faithfulness} in the context of machine learning explanations, both from the traditional feature importance perspective \cite{jacovi2020towards} and more recently for chain-of-thought reasoning for large language models \cite{agarwal2024faithfulness}. In this work, we focus on developing a technical framework for listener-adaptive explanations, and we do not make claims about the inner mechanism of a model based on the claims included in the utterances. Our approach can be coupled with existing methods that identify important concepts for the predictions of a black-box model in order to restrict the space of utterances allowed at each input. For example, one can test the statistical importance of each concept in the vocabulary \cite{teneggi2024testing} and only use significant ones to compose utterances.

Second, we remark that speakers are conditioned on the pretrained feature embeddings from the base classifier, meaning that they can generate utterances only using the information already contained in them. For example, if the embeddings were independent of color, then claims about color would appear randomly in the utterances, with no utility to a listener. \cref{fig:correlation} shows the correlation between the class-wise accuracy of the base classifier and of literal or pragmatic listener models. We report $R^2$ and Spearman's rank correlation coefficient $\rho$ as a function of minimal classifier accuracy. That is, for an accuracy threshold $\eta$, we consider all ground-truth classes where the accuracy of the base classifier is greater than or equal to $\eta$. Then, for each class, we regress the listener accuracy (\ie, whether it correctly inferred the predicted label) onto the classifier accuracy (\ie, whether it correctly predicted the class). We stratify results by listener type and utterance length. We found that pragmatic listeners (both with and without a warm-up phase in training) consistently correlate more with the base classifier than literal listeners across all accuracy thresholds on both datasets. This finding is interesting as neither the speaker nor the listener sees ground-truth labels during training, but predictions only. This suggests that jointly training the speaker and the listener in a pragmatic signaling game reflects the accuracy of the base classifier. In particular, pragmatic explanations for correct predictions have better communicative ability. Furthermore, we discuss the diversity of the generated utterances in \cref{supp:diversity}.

\subsection{\label{sec:adaptation}Pragmatic explanations adapt to listener preferences}
Here, we verify that our training procedure can successfully adapt utterances to different listeners via preference supervision only. In particular, we follow \cref{eq:temperature_scaling} and simulate listeners with different preferences by specifying prior distributions over groups of claims. Intuitively, as the distribution of topics in an utterance deviates from the prior, the output of the model becomes more uniform, making it harder to learn, and the strength of this effect can be increased with a temperature scale $\tau \geq 0$ such that when $\tau = 0$, the listener has no preference.

We study adaptation on the CUB and CheXpert datasets. For CUB---which includes annotations of 312 concepts describing various features of a bird---we categorize claims into 6 topics: bill features (29 claims), tail features (43 claims), head features (59 claims), color (129 claims), patterns (16 claims), and shape (24 claims). We consider two different listeners with their respective topic priors: a \emph{uniform} prior over all topics, and a \emph{topic} prior that includes claims about head, color, and patterns only. That is, the topic prior excludes claims about the bill, tail, and shape of a bird. This choice is motivated by the fact that these features may be unintuitive or difficult to distinguish, for example, because of occlusions in an image. 

For the CheXpert dataset, we consulted with an expert radiologist, and we augmented the original vocabulary of 12 medical findings by aggregating similar terms and describing them in layman's terms, for a total of 20 claims. For example, we aggregate ``enlarged cardiomediastinum'' with ``cardiomegaly'' into ``larger heart''; and ``pleural effusion'' with ``pleural other'' into ``fluid outside lung'' (see \cref{fig:chexpert_vocabulary} for the complete aggregation schema). In this case, we consider two listeners: a \emph{doctor} listener who prefers medical terms, and a \emph{patient} listener who favors layman's descriptions. We remark that concepts are represented as individual tokens in the speaker's vocabulary, so we make sure there are different numbers of laymen and medical terms, such that adaptation is not equivalent to finding a one-to-one mapping of the tokens.

\cref{fig:adaptation_kl} includes utterance adaptation results as a function of temperature scale. We evaluate adaptation in terms of the KL divergence between the empirical distribution of the groups of claims in the utterances generated with a speaker, and the target listener's prior (\ie, $\KL(g(u) || \pi)$). We normalize the KL divergence to a pragmatic speaker with no preference adaptation ($\tau = 0$). We can see that for all datasets and priors, a temperature parameter greater than 0 leads to a significant reduction in KL divergence. That is, speaker models successfully uncover and align with the simulated prior. For the CUB dataset, increasing the temperature scale further improves adaptation, whereas for the CheXpert dataset, values seem to plateau. Importantly, adaptation does not decrease listener accuracy (dashed lines in the figure). These results confirm that DPO is an effective technique to compose concept-based explanations that are communicative while adapting to unknown listener preferences.

\cref{fig:adaptation_example} includes some example comparisons of utterances generated with pragmatic speakers with and without preference adaptation (more examples are included in Appendix~\ref{supp:figures}). For CheXpert, we can see that a pragmatic speaker with no adaptation generates utterances that mix both medical and layman's terms (orange and blue, respectively). On the other hand, pragmatic speakers targeting doctor or patient listeners successfully concentrate on the correct type of claims. We can see that adapted utterances are not merely a translation of the same claims, highlighting that adaptation is more than a one-to-one mapping. For CUB, we show examples for a listener with the topic prior that excludes claims about the bill, tail, and shape of a bird. We highlight claims with colors according to their topic, where black includes topics excluded from the prior. As expected, adapted speakers generate utterances that match the desired prior.


\begin{figure}[t]
    \centering
    \includegraphics[width=0.9\linewidth]{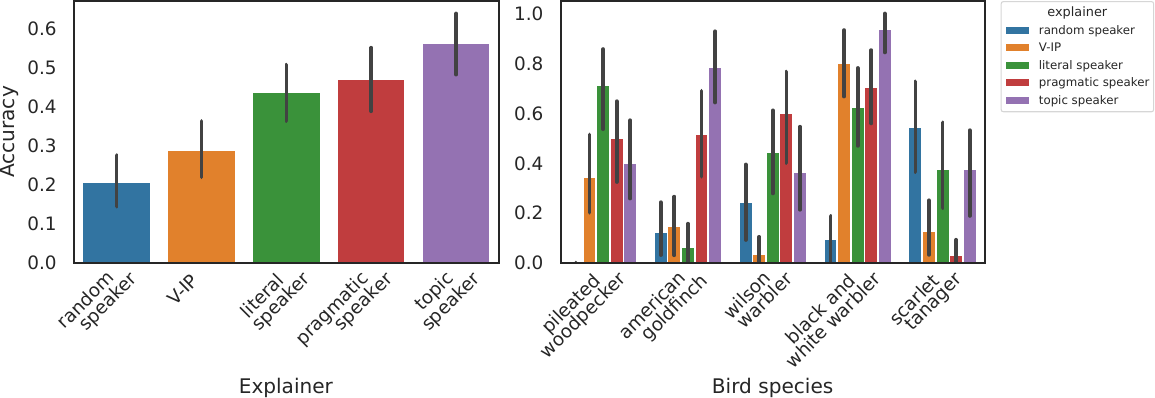}
    \caption{\label{fig:user_study_accuracy}Summary of results of our user study on the CUB dataset, both marginally per explainer and stratified by bird species. ``Topic speaker'' refers to a pragmatic speaker trained with a listener with a topic preference prior over claims about the head, coloration, and patterns of a bird.}
    \vspace{-10pt}
\end{figure}

\subsection{\label{sec:human_evaluation}User study}
In our experiments, so far, we have studied the behavior of our proposed algorithm with simulated listener models. But do pragmatic explanations improve the classification accuracy of humans? Here, we perform a user study to investigate whether good classification accuracy of simulated listener models transfers to real-world users of a machine learning model.\footnote{This study was approved by an Institutional Review Board (IRB).} This is important to validate that incorporating principles of pragmatic reasoning in machine learning interpretability methods leads to real-world improvements in the ability to communicate findings to different audiences.

We use the CUB dataset and design a survey where participants are asked to identify the species of different birds without seeing images, but only utterances. We compare a literal speaker, a pragmatic speaker with no adaptation, and a pragmatic speaker adapted to the same topic prior as in \cref{sec:adaptation} (\ie, features about the head, color, and patterns only). As baselines, we include a random speaker and V-IP. Recall that the CUB dataset contains 200 different bird species, and it would be impractical to expect participants to know the appearance of all of them. Therefore, we restrict the survey to 5 bird species to limit cognitive load. Furthermore, before the assessment, we introduce the task of bird identification using key features, present a slideshow of the bird species used in the study, and verify that all participants can solve a simple pre-assessment quiz, necessary to proceed to the assessment. During the assessment, participants are asked to identify the species of birds from a random subset of 4 out of 5 species for all conditions, for a total of 20 trials. We refer interested readers to Appendix~\ref{supp:user_study_flowchart} for a complete flowchart of the user study. Since the goal of this study is to verify that good accuracy of a listener model implies good accuracy of real-world users, we chose the top-5 bird species according to the mean accuracy of the two pragmatic listeners (with and without adaptation). Beyond accuracy, we chose species with balanced unique and shared traits in order to include ambiguities in the classification task---as is common in pragmatic reference expression generation \cite{cohn2018pragmatically}. For example, we include two yellow bird species: the American goldfinch and the Wilson warbler. Similarly, we include both the black and white warbler and the pileated woodpecker because they share black and white features (see Appendix~\ref{supp:user_study_species} for details on the selected bird species). We note that trials are randomized to remove bias and that we limit the assessment to 20 trials to control the duration of the assessment.

We recruited 40 participants between the ages of 18 and 65, whose primary language was English, and without colorblindness (see \cref{supp:user_study_cohort} for details on the cohort).\footnote{We recruited participants using Prolific \url{https://www.prolific.com/}.} \cref{fig:user_study_accuracy} summarizes the results of the study. First, we verify that utterances generated at random are uninformative and that, on average, participants guess one of the 5 bird species. Second, utterances generated with the adapted speaker are overall more communicative, with user accuracy of approximately $55\%$. This finding confirms that pragmatic reasoning is beneficial for composing utterances that are communicative in real-world scenarios. Furthermore, the fact that the utterances generated with the adapted speaker are more communicative than those with the unadapted pragmatic speaker suggests that the participants of our study align well with the topic prior. This highlights that direct preference optimization is an effective technical framework to adapt explanations to different listeners.

We stratify results by bird species, where no method consistently outperforms all others. For example, both V-IP and the adapted speaker achieve good classification performance for the black and white warbler, whereas for the scarlet tanager, all methods seem to struggle. We include a detailed discussion of these findings with example utterances and confusion matrices in Appendix~\ref{supp:user_study_results}.

\section{Conclusions and Limitations}
Being able to provide justifications for the predictions of complex machine learning models is increasingly important. While the literature on this question is extensive, most existing methods so far ignore the important aspects of contextual reasoning of the users who interact with these systems. This work proposes a framework to explain predictions of black-box models grounded in principles of pragmatic reasoning, the rational speech act, and direct preference optimization, as a step forward to address this gap. We provide extensive empirical evidence that the pragmatics of explanations are important for real-world applications with several different stakeholders. Importantly, we showed that:
\begin{itemize}[leftmargin=*]
    \item Employing ideas of pragmatic reasoning leads to speaker models that generate explanations that are more effective at communicating the prediction of the model being studied.
    \item Our framework generates explanations that adapt to different listeners, considering, for example, their levels of expertise or background knowledge.
    \item Increased utility of pragmatic explanations is reflected by the ability of participants in our user study to correctly infer the predicted class with better accuracy.
\end{itemize}
Overall, these results suggest that preference pairs from human feedback and DPO are a promising technical framework for practical scenarios where it is difficult to model listeners directly. Moving forward, we hope our techniques enable the deployment of explanations of machine learning models that best serve a particular population of humans.

Our work still has limitations. First, our training procedure relies on iterative refinement of the speaker and listener models, which may be expensive for large datasets (see \cref{table:complexity} for runtimes). In particular, the space of utterances grows exponentially with the number of claims in the vocabulary $\C$, leading to an exploration-exploitation tradeoff when sampling preference and explanation datasets in the training procedure. In our experiments, we studied the effects of adding a random warm-up phase to prevent speakers and listeners from converging to suboptimal solutions. Other strategies to increase coverage of the space of utterances exist, such as retraining candidate utterances from previous iterations \cite{rosset2024direct}, or using an exponential moving average for the reference model \cite{meta2024llama}.

Second, free-form explanations may be more expressive than concept-based ones, better representing the way we communicate. This approach requires claim decomposition and verification to compute consistency and prevent hallucinations \cite{kamoi2023wice,wanner2024dndscore}, which may lead to deceitful explanations.

Finally, training speaker models that can adapt to multiple listeners at once could be a useful extension and is a matter of future work.

\subsection*{Acknowledgments}
This research was supported by NSF CAREER Award CCF 2239787.

\newpage
\clearpage
\bibliographystyle{plainnat}
\bibliography{bibliography}

\begin{thebibliography}{121}
\providecommand{\natexlab}[1]{#1}
\providecommand{\url}[1]{\texttt{#1}}
\expandafter\ifx\csname urlstyle\endcsname\relax
  \providecommand{\doi}[1]{doi: #1}\else
  \providecommand{\doi}{doi: \begingroup \urlstyle{rm}\Url}\fi

\bibitem[Agarwal et~al.(2024)Agarwal, Tanneru, and Lakkaraju]{agarwal2024faithfulness}
Chirag Agarwal, Sree~Harsha Tanneru, and Himabindu Lakkaraju.
\newblock Faithfulness vs. plausibility: On the (un) reliability of explanations from large language models.
\newblock \emph{arXiv preprint arXiv:2402.04614}, 2024.

\bibitem[Andreas and Klein(2016)]{andreas2016reasoning}
Jacob Andreas and Dan Klein.
\newblock Reasoning about pragmatics with neural listeners and speakers.
\newblock \emph{arXiv preprint arXiv:1604.00562}, 2016.

\bibitem[Askell et~al.(2021)Askell, Bai, Chen, Drain, Ganguli, Henighan, Jones, Joseph, Mann, DasSarma, et~al.]{askell2021general}
Amanda Askell, Yuntao Bai, Anna Chen, Dawn Drain, Deep Ganguli, Tom Henighan, Andy Jones, Nicholas Joseph, Ben Mann, Nova DasSarma, et~al.
\newblock A general language assistant as a laboratory for alignment.
\newblock \emph{arXiv preprint arXiv:2112.00861}, 2021.

\bibitem[Baker et~al.(2009)Baker, Saxe, and Tenenbaum]{baker2009action}
Chris~L Baker, Rebecca Saxe, and Joshua~B Tenenbaum.
\newblock Action understanding as inverse planning.
\newblock \emph{Cognition}, 113\penalty0 (3):\penalty0 329--349, 2009.

\bibitem[Bannur et~al.(2023)Bannur, Hyland, Liu, Perez-Garcia, Ilse, Castro, Boecking, Sharma, Bouzid, Thieme, et~al.]{bannur2023learning}
Shruthi Bannur, Stephanie Hyland, Qianchu Liu, Fernando Perez-Garcia, Maximilian Ilse, Daniel~C Castro, Benedikt Boecking, Harshita Sharma, Kenza Bouzid, Anja Thieme, et~al.
\newblock Learning to exploit temporal structure for biomedical vision-language processing.
\newblock In \emph{Proceedings of the IEEE/CVF Conference on Computer Vision and Pattern Recognition}, pages 15016--15027, 2023.

\bibitem[Bereska and Gavves(2024)]{bereska2024mechanistic}
Leonard Bereska and Efstratios Gavves.
\newblock Mechanistic interpretability for ai safety--a review.
\newblock \emph{arXiv preprint arXiv:2404.14082}, 2024.

\bibitem[Bertrand et~al.(2023)Bertrand, Viard, Belloum, Eagan, and Maxwell]{bertrand2023selective}
Astrid Bertrand, Tiphaine Viard, Rafik Belloum, James~R Eagan, and Winston Maxwell.
\newblock On selective, mutable and dialogic xai: A review of what users say about different types of interactive explanations.
\newblock In \emph{Proceedings of the 2023 CHI Conference on Human Factors in Computing Systems}, pages 1--21, 2023.

\bibitem[Bharti et~al.(2024)Bharti, Yi, and Sulam]{bharti2024sufficient}
Beepul Bharti, Paul Yi, and Jeremias Sulam.
\newblock Sufficient and necessary explanations (and what lies in between).
\newblock \emph{arXiv preprint arXiv:2409.20427}, 2024.

\bibitem[Bradley and Terry(1952)]{bradley1952rank}
Ralph~Allan Bradley and Milton~E Terry.
\newblock Rank analysis of incomplete block designs: I. the method of paired comparisons.
\newblock \emph{Biometrika}, 39\penalty0 (3/4):\penalty0 324--345, 1952.

\bibitem[Brandizzi(2023)]{brandizzi2023toward}
Nicolo’ Brandizzi.
\newblock Toward more human-like ai communication: A review of emergent communication research.
\newblock \emph{IEEE Access}, 11:\penalty0 142317--142340, 2023.

\bibitem[Bromberger(1984)]{bromberger1984pragmatic}
Sylvai Bromberger.
\newblock On pragmatic and scientific explanation: Comments on achinstein's and salmon's papers.
\newblock In \emph{PSA: Proceedings of the Biennial Meeting of the Philosophy of Science Association}, volume 1984, pages 306--325. Cambridge University Press, 1984.

\bibitem[Bromberger(1992)]{bromberger1992we}
Sylvain Bromberger.
\newblock \emph{On what we know we don't know: Explanation, theory, linguistics, and how questions shape them}.
\newblock University of Chicago Press, 1992.

\bibitem[Bubeck et~al.(2023)Bubeck, Chadrasekaran, Eldan, Gehrke, Horvitz, Kamar, Lee, Lee, Li, Lundberg, et~al.]{bubeck2023sparks}
S{\'e}bastien Bubeck, Varun Chadrasekaran, Ronen Eldan, Johannes Gehrke, Eric Horvitz, Ece Kamar, Peter Lee, Yin~Tat Lee, Yuanzhi Li, Scott Lundberg, et~al.
\newblock Sparks of artificial general intelligence: Early experiments with gpt-4, 2023.

\bibitem[Burns et~al.(2020)Burns, Thomason, and Tansey]{burns2020interpreting}
Collin Burns, Jesse Thomason, and Wesley Tansey.
\newblock Interpreting black box models via hypothesis testing.
\newblock In \emph{Proceedings of the 2020 ACM-IMS on foundations of data science conference}, pages 47--57, 2020.

\bibitem[Carston et~al.(1993)]{carston1993conjunction}
Robyn Carston et~al.
\newblock Conjunction, explanation and relevance.
\newblock 1993.

\bibitem[Charoenphakdee et~al.(2021)Charoenphakdee, Cui, Zhang, and Sugiyama]{charoenphakdee2021classification}
Nontawat Charoenphakdee, Zhenghang Cui, Yivan Zhang, and Masashi Sugiyama.
\newblock Classification with rejection based on cost-sensitive classification.
\newblock In \emph{International Conference on Machine Learning}, pages 1507--1517. PMLR, 2021.

\bibitem[Chattopadhyay et~al.(2023)Chattopadhyay, Chan, Haeffele, Geman, and Vidal]{chattopadhyay2023variational}
Aditya Chattopadhyay, Kwan Ho~Ryan Chan, Benjamin~D Haeffele, Donald Geman, and Rene Vidal.
\newblock Variational information pursuit for interpretable predictions.
\newblock \emph{arXiv preprint arXiv:2302.02876}, 2023.

\bibitem[Chattopadhyay et~al.(2024)Chattopadhyay, Chan, and Vidal]{chattopadhyay2024bootstrapping}
Aditya Chattopadhyay, Kwan Ho~Ryan Chan, and Rene Vidal.
\newblock Bootstrapping variational information pursuit with large language and vision models for interpretable image classification.
\newblock In \emph{The Twelfth International Conference on Learning Representations}, 2024.

\bibitem[Cherti et~al.(2023)Cherti, Beaumont, Wightman, Wortsman, Ilharco, Gordon, Schuhmann, Schmidt, and Jitsev]{cherti2023reproducible}
Mehdi Cherti, Romain Beaumont, Ross Wightman, Mitchell Wortsman, Gabriel Ilharco, Cade Gordon, Christoph Schuhmann, Ludwig Schmidt, and Jenia Jitsev.
\newblock Reproducible scaling laws for contrastive language-image learning.
\newblock In \emph{Proceedings of the IEEE/CVF conference on computer vision and pattern recognition}, pages 2818--2829, 2023.

\bibitem[Cohn-Gordon et~al.(2018)Cohn-Gordon, Goodman, and Potts]{cohn2018pragmatically}
Reuben Cohn-Gordon, Noah Goodman, and Christopher Potts.
\newblock Pragmatically informative image captioning with character-level inference.
\newblock \emph{arXiv preprint arXiv:1804.05417}, 2018.

\bibitem[Conmy et~al.(2023)Conmy, Mavor-Parker, Lynch, Heimersheim, and Garriga-Alonso]{conmy2023towards}
Arthur Conmy, Augustine Mavor-Parker, Aengus Lynch, Stefan Heimersheim, and Adri{\`a} Garriga-Alonso.
\newblock Towards automated circuit discovery for mechanistic interpretability.
\newblock \emph{Advances in Neural Information Processing Systems}, 36:\penalty0 16318--16352, 2023.

\bibitem[Cortes et~al.(2016)Cortes, DeSalvo, and Mohri]{cortes2016learning}
Corinna Cortes, Giulia DeSalvo, and Mehryar Mohri.
\newblock Learning with rejection.
\newblock In \emph{International conference on algorithmic learning theory}, pages 67--82. Springer, 2016.

\bibitem[Corti et~al.(2024)Corti, Oltmans, Jung, Balayn, Wijsenbeek, and Yang]{corti2024moving}
Lorenzo Corti, Rembrandt Oltmans, Jiwon Jung, Agathe Balayn, Marlies Wijsenbeek, and Jie Yang.
\newblock ``it is a moving process": Understanding the evolution of explainability needs of clinicians in pulmonary medicine.
\newblock In \emph{Proceedings of the 2024 CHI Conference on Human Factors in Computing Systems}, pages 1--21, 2024.

\bibitem[Covert et~al.(2020)Covert, Lundberg, and Lee]{covert2020understanding}
Ian Covert, Scott~M Lundberg, and Su-In Lee.
\newblock Understanding global feature contributions with additive importance measures.
\newblock \emph{Advances in Neural Information Processing Systems}, 33:\penalty0 17212--17223, 2020.

\bibitem[Covert et~al.(2021)Covert, Lundberg, and Lee]{covert2021explaining}
Ian Covert, Scott Lundberg, and Su-In Lee.
\newblock Explaining by removing: A unified framework for model explanation.
\newblock \emph{Journal of Machine Learning Research}, 22\penalty0 (209):\penalty0 1--90, 2021.

\bibitem[Cunningham et~al.(2023)Cunningham, Ewart, Riggs, Huben, and Sharkey]{cunningham2023sparse}
Hoagy Cunningham, Aidan Ewart, Logan Riggs, Robert Huben, and Lee Sharkey.
\newblock Sparse autoencoders find highly interpretable features in language models.
\newblock \emph{arXiv preprint arXiv:2309.08600}, 2023.

\bibitem[Daneshjou et~al.(2022)Daneshjou, Yuksekgonul, Cai, Novoa, and Zou]{daneshjou2022skincon}
Roxana Daneshjou, Mert Yuksekgonul, Zhuo~Ran Cai, Roberto Novoa, and James~Y Zou.
\newblock Skincon: A skin disease dataset densely annotated by domain experts for fine-grained debugging and analysis.
\newblock \emph{Advances in Neural Information Processing Systems}, 35:\penalty0 18157--18167, 2022.

\bibitem[De~Leeuw(2015)]{de2015jspsych}
Joshua~R De~Leeuw.
\newblock jspsych: A javascript library for creating behavioral experiments in a web browser.
\newblock \emph{Behavior research methods}, 47:\penalty0 1--12, 2015.

\bibitem[De~Regt(2017)]{de2017understanding}
Henk~W De~Regt.
\newblock \emph{Understanding scientific understanding}.
\newblock Oxford university press, 2017.

\bibitem[Deng et~al.(2009)Deng, Dong, Socher, Li, Li, and Fei-Fei]{deng2009imagenet}
Jia Deng, Wei Dong, Richard Socher, Li-Jia Li, Kai Li, and Li~Fei-Fei.
\newblock Imagenet: A large-scale hierarchical image database.
\newblock In \emph{2009 IEEE conference on computer vision and pattern recognition}, pages 248--255. Ieee, 2009.

\bibitem[Devlin et~al.(2019)Devlin, Chang, Lee, and Toutanova]{devlin2019bert}
Jacob Devlin, Ming-Wei Chang, Kenton Lee, and Kristina Toutanova.
\newblock Bert: Pre-training of deep bidirectional transformers for language understanding.
\newblock In \emph{Proceedings of the 2019 conference of the North American chapter of the association for computational linguistics: human language technologies, volume 1 (long and short papers)}, pages 4171--4186, 2019.

\bibitem[Dunefsky et~al.(2024)Dunefsky, Chlenski, and Nanda]{dunefsky2024transcoders}
Jacob Dunefsky, Philippe Chlenski, and Neel Nanda.
\newblock Transcoders find interpretable llm feature circuits.
\newblock \emph{Advances in Neural Information Processing Systems}, 37:\penalty0 24375--24410, 2024.

\bibitem[Ehsan et~al.(2021)Ehsan, Liao, Muller, Riedl, and Weisz]{ehsan2021expanding}
Upol Ehsan, Q~Vera Liao, Michael Muller, Mark~O Riedl, and Justin~D Weisz.
\newblock Expanding explainability: Towards social transparency in ai systems.
\newblock In \emph{Proceedings of the 2021 CHI conference on human factors in computing systems}, pages 1--19, 2021.

\bibitem[Frank and Goodman(2012)]{frank2012predicting}
Michael~C Frank and Noah~D Goodman.
\newblock Predicting pragmatic reasoning in language games.
\newblock \emph{Science}, 336\penalty0 (6084):\penalty0 998--998, 2012.

\bibitem[Gabriel(2020)]{gabriel2020artificial}
Iason Gabriel.
\newblock Artificial intelligence, values, and alignment.
\newblock \emph{Minds and machines}, 30\penalty0 (3):\penalty0 411--437, 2020.

\bibitem[Gibbs~Jr and Moise(1997)]{gibbs1997pragmatics}
Raymond~W Gibbs~Jr and Jessica~F Moise.
\newblock Pragmatics in understanding what is said.
\newblock \emph{Cognition}, 62\penalty0 (1):\penalty0 51--74, 1997.

\bibitem[Goddu and Gopnik(2024)]{goddu2024development}
Mariel~K Goddu and Alison Gopnik.
\newblock The development of human causal learning and reasoning.
\newblock \emph{Nature Reviews Psychology}, 3\penalty0 (5):\penalty0 319--339, 2024.

\bibitem[Goodman and Frank(2016)]{goodman2016pragmatic}
Noah~D Goodman and Michael~C Frank.
\newblock Pragmatic language interpretation as probabilistic inference.
\newblock \emph{Trends in cognitive sciences}, 20\penalty0 (11):\penalty0 818--829, 2016.

\bibitem[Greco et~al.(2023)Greco, Bagade, Le, and Bernardi]{greco2023she}
Claudio Greco, Diksha Bagade, Dieu-Thu Le, and Raffaella Bernardi.
\newblock She adapts to her student: An expert pragmatic speaker tailoring her referring expressions to the layman listener.
\newblock \emph{Frontiers in Artificial Intelligence}, 6:\penalty0 1017204, 2023.

\bibitem[Grice(1975)]{grice1975logic}
Herbert~P Grice.
\newblock Logic and conversation.
\newblock In \emph{Speech acts}, pages 41--58. Brill, 1975.

\bibitem[Guidotti(2024)]{guidotti2024counterfactual}
Riccardo Guidotti.
\newblock Counterfactual explanations and how to find them: literature review and benchmarking.
\newblock \emph{Data Mining and Knowledge Discovery}, 38\penalty0 (5):\penalty0 2770--2824, 2024.

\bibitem[Guo et~al.(2017)Guo, Pleiss, Sun, and Weinberger]{guo2017calibration}
Chuan Guo, Geoff Pleiss, Yu~Sun, and Kilian~Q Weinberger.
\newblock On calibration of modern neural networks.
\newblock In \emph{International conference on machine learning}, pages 1321--1330. PMLR, 2017.

\bibitem[Hadfield-Menell et~al.(2016)Hadfield-Menell, Russell, Abbeel, and Dragan]{hadfield2016cooperative}
Dylan Hadfield-Menell, Stuart~J Russell, Pieter Abbeel, and Anca Dragan.
\newblock Cooperative inverse reinforcement learning.
\newblock \emph{Advances in neural information processing systems}, 29, 2016.

\bibitem[Han et~al.(2023)Han, Ektefaie, Farhat, Zitnik, and Lakkaraju]{han2023ignorance}
Tessa Han, Yasha Ektefaie, Maha Farhat, Marinka Zitnik, and Himabindu Lakkaraju.
\newblock Is ignorance bliss? the role of post hoc explanation faithfulness and alignment in model trust in laypeople and domain experts.
\newblock \emph{arXiv preprint arXiv:2312.05690}, 2023.

\bibitem[Haque et~al.(2023)Haque, Islam, and Mikalef]{haque2023explainable}
AKM~Bahalul Haque, AKM~Najmul Islam, and Patrick Mikalef.
\newblock Explainable artificial intelligence (xai) from a user perspective: A synthesis of prior literature and problematizing avenues for future research.
\newblock \emph{Technological Forecasting and Social Change}, 186:\penalty0 122120, 2023.

\bibitem[Harding et~al.(2025)Harding, Gerstenberg, and Icard]{harding2025communication}
Jacqueline Harding, Tobias Gerstenberg, and Thomas Icard.
\newblock A communication-first account of explanation.
\newblock \emph{arXiv preprint arXiv:2505.03732}, 2025.

\bibitem[Hawkins et~al.(2019)Hawkins, Kwon, Sadigh, and Goodman]{hawkins2019continual}
Robert~D Hawkins, Minae Kwon, Dorsa Sadigh, and Noah~D Goodman.
\newblock Continual adaptation for efficient machine communication.
\newblock \emph{arXiv preprint arXiv:1911.09896}, 2019.

\bibitem[Hendrickx et~al.(2024)Hendrickx, Perini, Van~der Plas, Meert, and Davis]{hendrickx2024machine}
Kilian Hendrickx, Lorenzo Perini, Dries Van~der Plas, Wannes Meert, and Jesse Davis.
\newblock Machine learning with a reject option: A survey.
\newblock \emph{Machine Learning}, 113\penalty0 (5):\penalty0 3073--3110, 2024.

\bibitem[Hobbs et~al.(1987)]{hobbs1987implicature}
Jerry~R Hobbs et~al.
\newblock \emph{Implicature and definite reference}.
\newblock CSLI, 1987.

\bibitem[Irvin et~al.(2019)Irvin, Rajpurkar, Ko, Yu, Ciurea-Ilcus, Chute, Marklund, Haghgoo, Ball, Shpanskaya, et~al.]{irvin2019chexpert}
Jeremy Irvin, Pranav Rajpurkar, Michael Ko, Yifan Yu, Silviana Ciurea-Ilcus, Chris Chute, Henrik Marklund, Behzad Haghgoo, Robyn Ball, Katie Shpanskaya, et~al.
\newblock Chexpert: A large chest radiograph dataset with uncertainty labels and expert comparison.
\newblock In \emph{Proceedings of the AAAI conference on artificial intelligence}, volume~33, pages 590--597, 2019.

\bibitem[Jacovi and Goldberg(2020)]{jacovi2020towards}
Alon Jacovi and Yoav Goldberg.
\newblock Towards faithfully interpretable nlp systems: How should we define and evaluate faithfulness?
\newblock \emph{arXiv preprint arXiv:2004.03685}, 2020.

\bibitem[Jain et~al.(2021)Jain, Smit, Truong, Nguyen, Huynh, Jain, Young, Ng, Lungren, and Rajpurkar]{jain2021visualchexbert}
Saahil Jain, Akshay Smit, Steven~QH Truong, Chanh~DT Nguyen, Minh-Thanh Huynh, Mudit Jain, Victoria~A Young, Andrew~Y Ng, Matthew~P Lungren, and Pranav Rajpurkar.
\newblock Visualchexbert: addressing the discrepancy between radiology report labels and image labels.
\newblock In \emph{Proceedings of the Conference on Health, Inference, and Learning}, pages 105--115, 2021.

\bibitem[Jara-Ettinger et~al.(2016)Jara-Ettinger, Gweon, Schulz, and Tenenbaum]{jara2016naive}
Julian Jara-Ettinger, Hyowon Gweon, Laura~E Schulz, and Joshua~B Tenenbaum.
\newblock The na{\"\i}ve utility calculus: Computational principles underlying commonsense psychology.
\newblock \emph{Trends in cognitive sciences}, 20\penalty0 (8):\penalty0 589--604, 2016.

\bibitem[Jiang et~al.(2018)Jiang, Kim, Guan, and Gupta]{jiang2018trust}
Heinrich Jiang, Been Kim, Melody Guan, and Maya Gupta.
\newblock To trust or not to trust a classifier.
\newblock \emph{Advances in neural information processing systems}, 31, 2018.

\bibitem[Kamoi et~al.(2023)Kamoi, Goyal, Rodriguez, and Durrett]{kamoi2023wice}
Ryo Kamoi, Tanya Goyal, Juan~Diego Rodriguez, and Greg Durrett.
\newblock Wice: Real-world entailment for claims in wikipedia.
\newblock \emph{arXiv preprint arXiv:2303.01432}, 2023.

\bibitem[Kaufmann et~al.(2023)Kaufmann, Weng, Bengs, and H{\"u}llermeier]{kaufmann2023survey}
Timo Kaufmann, Paul Weng, Viktor Bengs, and Eyke H{\"u}llermeier.
\newblock A survey of reinforcement learning from human feedback.
\newblock \emph{arXiv preprint arXiv:2312.14925}, 10, 2023.

\bibitem[Kaur et~al.(2020)Kaur, Nori, Jenkins, Caruana, Wallach, and Wortman~Vaughan]{kaur2020interpreting}
Harmanpreet Kaur, Harsha Nori, Samuel Jenkins, Rich Caruana, Hanna Wallach, and Jennifer Wortman~Vaughan.
\newblock Interpreting interpretability: understanding data scientists' use of interpretability tools for machine learning.
\newblock In \emph{Proceedings of the 2020 CHI conference on human factors in computing systems}, pages 1--14, 2020.

\bibitem[Kim et~al.(2018)Kim, Wattenberg, Gilmer, Cai, Wexler, Viegas, et~al.]{kim2018interpretability}
Been Kim, Martin Wattenberg, Justin Gilmer, Carrie Cai, James Wexler, Fernanda Viegas, et~al.
\newblock Interpretability beyond feature attribution: Quantitative testing with concept activation vectors (tcav).
\newblock In \emph{International conference on machine learning}, pages 2668--2677. PMLR, 2018.

\bibitem[Kim et~al.(2025)Kim, Hewitt, Nanda, Fiedel, and Tafjord]{kim2025because}
Been Kim, John Hewitt, Neel Nanda, Noah Fiedel, and Oyvind Tafjord.
\newblock Because we have llms, we can and should pursue agentic interpretability.
\newblock \emph{arXiv preprint arXiv:2506.12152}, 2025.

\bibitem[Koh et~al.(2020)Koh, Nguyen, Tang, Mussmann, Pierson, Kim, and Liang]{koh2020concept}
Pang~Wei Koh, Thao Nguyen, Yew~Siang Tang, Stephen Mussmann, Emma Pierson, Been Kim, and Percy Liang.
\newblock Concept bottleneck models.
\newblock In \emph{International conference on machine learning}, pages 5338--5348. PMLR, 2020.

\bibitem[Kolek et~al.(2020)Kolek, Nguyen, Levie, Bruna, and Kutyniok]{kolek2020rate}
Stefan Kolek, Duc~Anh Nguyen, Ron Levie, Joan Bruna, and Gitta Kutyniok.
\newblock A rate-distortion framework for explaining black-box model decisions.
\newblock In \emph{International Workshop on Extending Explainable AI Beyond Deep Models and Classifiers}, pages 91--115. Springer, 2020.

\bibitem[Kolek et~al.(2022)Kolek, Nguyen, Levie, Bruna, and Kutyniok]{kolek2022cartoon}
Stefan Kolek, Duc~Anh Nguyen, Ron Levie, Joan Bruna, and Gitta Kutyniok.
\newblock Cartoon explanations of image classifiers.
\newblock In \emph{European Conference on Computer Vision}, pages 443--458. Springer, 2022.

\bibitem[Krahmer and Van~Deemter(2012)]{krahmer2012computational}
Emiel Krahmer and Kees Van~Deemter.
\newblock Computational generation of referring expressions: A survey.
\newblock \emph{Computational Linguistics}, 38\penalty0 (1):\penalty0 173--218, 2012.

\bibitem[Lei et~al.(2016)Lei, Barzilay, and Jaakkola]{lei2016rationalizing}
Tao Lei, Regina Barzilay, and Tommi Jaakkola.
\newblock Rationalizing neural predictions.
\newblock \emph{arXiv preprint arXiv:1606.04155}, 2016.

\bibitem[Liu et~al.(2025)Liu, Fang, Hu, Zhang, Zhou, Zhang, Tu, Lin, Huang, Song, et~al.]{liu2025survey}
Shunyu Liu, Wenkai Fang, Zetian Hu, Junjie Zhang, Yang Zhou, Kongcheng Zhang, Rongcheng Tu, Ting-En Lin, Fei Huang, Mingli Song, et~al.
\newblock A survey of direct preference optimization.
\newblock \emph{arXiv preprint arXiv:2503.11701}, 2025.

\bibitem[Liu et~al.(2024)Liu, Deng, Niu, Wang, Wang, Zhang, and Li]{liu2024mmi}
Wei Liu, Zhiying Deng, Zhongyu Niu, Jun Wang, Haozhao Wang, YuanKai Zhang, and Ruixuan Li.
\newblock Is the mmi criterion necessary for interpretability? degenerating non-causal features to plain noise for self-rationalization.
\newblock \emph{Advances in Neural Information Processing Systems}, 37:\penalty0 117636--117656, 2024.

\bibitem[Llama~Team(2024)]{meta2024llama}
AI~@~Meta Llama~Team.
\newblock The llama 3 herd of models.
\newblock \emph{arXiv preprint arXiv:2407.21783}, 2024.

\bibitem[Loshchilov and Hutter(2016)]{loshchilov2016sgdr}
Ilya Loshchilov and Frank Hutter.
\newblock Sgdr: Stochastic gradient descent with warm restarts.
\newblock \emph{arXiv preprint arXiv:1608.03983}, 2016.

\bibitem[Loshchilov and Hutter(2017)]{loshchilov2017decoupled}
Ilya Loshchilov and Frank Hutter.
\newblock Decoupled weight decay regularization.
\newblock \emph{arXiv preprint arXiv:1711.05101}, 2017.

\bibitem[Lundberg and Lee(2017)]{lundberg2017unified}
Scott~M Lundberg and Su-In Lee.
\newblock A unified approach to interpreting model predictions.
\newblock \emph{Advances in neural information processing systems}, 30, 2017.

\bibitem[Miller(2019)]{miller2019explanation}
Tim Miller.
\newblock Explanation in artificial intelligence: Insights from the social sciences.
\newblock \emph{Artificial intelligence}, 267:\penalty0 1--38, 2019.

\bibitem[Montavon et~al.(2018)Montavon, Samek, and M{\"u}ller]{montavon2018methods}
Gr{\'e}goire Montavon, Wojciech Samek, and Klaus-Robert M{\"u}ller.
\newblock Methods for interpreting and understanding deep neural networks.
\newblock \emph{Digital signal processing}, 73:\penalty0 1--15, 2018.

\bibitem[Nanda et~al.(2023)Nanda, Chan, Lieberum, Smith, and Steinhardt]{nanda2023progress}
Neel Nanda, Lawrence Chan, Tom Lieberum, Jess Smith, and Jacob Steinhardt.
\newblock Progress measures for grokking via mechanistic interpretability.
\newblock \emph{arXiv preprint arXiv:2301.05217}, 2023.

\bibitem[Nyrup and Robinson(2022)]{nyrup2022explanatory}
Rune Nyrup and Diana Robinson.
\newblock Explanatory pragmatism: a context-sensitive framework for explainable medical ai.
\newblock \emph{Ethics and information technology}, 24\penalty0 (1):\penalty0 13, 2022.

\bibitem[Ou et~al.(2023)Ou, Krojer, and Fried]{ou2023pragmatic}
Jiefu Ou, Benno Krojer, and Daniel Fried.
\newblock Pragmatic inference with a clip listener for contrastive captioning.
\newblock \emph{arXiv preprint arXiv:2306.08818}, 2023.

\bibitem[Ouyang et~al.(2022)Ouyang, Wu, Jiang, Almeida, Wainwright, Mishkin, Zhang, Agarwal, Slama, Ray, et~al.]{ouyang2022training}
Long Ouyang, Jeffrey Wu, Xu~Jiang, Diogo Almeida, Carroll Wainwright, Pamela Mishkin, Chong Zhang, Sandhini Agarwal, Katarina Slama, Alex Ray, et~al.
\newblock Training language models to follow instructions with human feedback.
\newblock \emph{Advances in neural information processing systems}, 35:\penalty0 27730--27744, 2022.

\bibitem[P{\'a}ez(2019)]{paez2019pragmatic}
Andr{\'e}s P{\'a}ez.
\newblock The pragmatic turn in explainable artificial intelligence (xai).
\newblock \emph{Minds and Machines}, 29\penalty0 (3):\penalty0 441--459, 2019.

\bibitem[Pang et~al.(2024)Pang, Yuan, He, Cho, Sukhbaatar, and Weston]{pang2024iterative}
Richard~Yuanzhe Pang, Weizhe Yuan, He~He, Kyunghyun Cho, Sainbayar Sukhbaatar, and Jason Weston.
\newblock Iterative reasoning preference optimization.
\newblock \emph{Advances in Neural Information Processing Systems}, 37:\penalty0 116617--116637, 2024.

\bibitem[Panigrahi et~al.(2025)Panigrahi, Kim, Liaqat, Jinturkar, Russakovsky, Fong, and Abtahi]{panigrahi2025interactivity}
Indu Panigrahi, Sunnie~SY Kim, Amna Liaqat, Rohan Jinturkar, Olga Russakovsky, Ruth Fong, and Parastoo Abtahi.
\newblock Interactivity x explainability: Toward understanding how interactivity can improve computer vision explanations.
\newblock \emph{arXiv preprint arXiv:2504.10745}, 2025.

\bibitem[Peng et~al.(2024)Peng, Sun, Shu, and Abel]{peng2024pragmatic}
Andi Peng, Yuying Sun, Tianmin Shu, and David Abel.
\newblock Pragmatic feature preferences: learning reward-relevant preferences from human input.
\newblock \emph{arXiv preprint arXiv:2405.14769}, 2024.

\bibitem[Premack and Woodruff(1978)]{premack1978does}
David Premack and Guy Woodruff.
\newblock Does the chimpanzee have a theory of mind?
\newblock \emph{Behavioral and brain sciences}, 1\penalty0 (4):\penalty0 515--526, 1978.

\bibitem[Pu et~al.(2024)Pu, Vaduguru, Vaithilingam, Glassman, and Fried]{pu2024amortizing}
Yewen Pu, Saujas Vaduguru, Priyan Vaithilingam, Elena Glassman, and Daniel Fried.
\newblock Amortizing pragmatic program synthesis with rankings.
\newblock \emph{arXiv preprint arXiv:2407.02499}, 2024.

\bibitem[Radford et~al.(2021)Radford, Kim, Hallacy, Ramesh, Goh, Agarwal, Sastry, Askell, Mishkin, Clark, et~al.]{radford2021learning}
Alec Radford, Jong~Wook Kim, Chris Hallacy, Aditya Ramesh, Gabriel Goh, Sandhini Agarwal, Girish Sastry, Amanda Askell, Pamela Mishkin, Jack Clark, et~al.
\newblock Learning transferable visual models from natural language supervision.
\newblock In \emph{International conference on machine learning}, pages 8748--8763. PmLR, 2021.

\bibitem[Rafailov et~al.(2023)Rafailov, Sharma, Mitchell, Manning, Ermon, and Finn]{rafailov2023direct}
Rafael Rafailov, Archit Sharma, Eric Mitchell, Christopher~D Manning, Stefano Ermon, and Chelsea Finn.
\newblock Direct preference optimization: Your language model is secretly a reward model.
\newblock \emph{Advances in Neural Information Processing Systems}, 36:\penalty0 53728--53741, 2023.

\bibitem[Recanati(1989)]{recanati1989pragmatics}
Fran{\c{c}}ois Recanati.
\newblock The pragmatics of what is said.
\newblock 1989.

\bibitem[Rosset et~al.(2024)Rosset, Cheng, Mitra, Santacroce, Awadallah, and Xie]{rosset2024direct}
Corby Rosset, Ching-An Cheng, Arindam Mitra, Michael Santacroce, Ahmed Awadallah, and Tengyang Xie.
\newblock Direct nash optimization: Teaching language models to self-improve with general preferences.
\newblock \emph{arXiv preprint arXiv:2404.03715}, 2024.

\bibitem[Russakovsky et~al.(2015)Russakovsky, Deng, Su, Krause, Satheesh, Ma, Huang, Karpathy, Khosla, Bernstein, et~al.]{russakovsky2015imagenet}
Olga Russakovsky, Jia Deng, Hao Su, Jonathan Krause, Sanjeev Satheesh, Sean Ma, Zhiheng Huang, Andrej Karpathy, Aditya Khosla, Michael Bernstein, et~al.
\newblock Imagenet large scale visual recognition challenge.
\newblock \emph{International journal of computer vision}, 115:\penalty0 211--252, 2015.

\bibitem[Russell(2019)]{russell2019human}
Stuart Russell.
\newblock \emph{Human compatible: AI and the problem of control}.
\newblock Penguin Uk, 2019.

\bibitem[Schlotterbeck and Wang(2023)]{schlotterbeck2023incremental}
Fabian Schlotterbeck and Hening Wang.
\newblock An incremental rsa model for adjective ordering preferences in referential visual context.
\newblock \emph{Society for Computation in Linguistics}, 6\penalty0 (1), 2023.

\bibitem[Schneider and Handali(2019)]{schneider2019personalized}
Johanes Schneider and Joshua Handali.
\newblock Personalized explanation in machine learning: A conceptualization.
\newblock \emph{arXiv preprint arXiv:1901.00770}, 2019.

\bibitem[Schoonderwoerd et~al.(2021)Schoonderwoerd, Jorritsma, Neerincx, and Van Den~Bosch]{schoonderwoerd2021human}
Tjeerd~AJ Schoonderwoerd, Wiard Jorritsma, Mark~A Neerincx, and Karel Van Den~Bosch.
\newblock Human-centered xai: Developing design patterns for explanations of clinical decision support systems.
\newblock \emph{International Journal of Human-Computer Studies}, 154:\penalty0 102684, 2021.

\bibitem[Schuhmann et~al.(2022)Schuhmann, Beaumont, Vencu, Gordon, Wightman, Cherti, Coombes, Katta, Mullis, Wortsman, et~al.]{schuhmann2022laion}
Christoph Schuhmann, Romain Beaumont, Richard Vencu, Cade Gordon, Ross Wightman, Mehdi Cherti, Theo Coombes, Aarush Katta, Clayton Mullis, Mitchell Wortsman, et~al.
\newblock Laion-5b: An open large-scale dataset for training next generation image-text models.
\newblock \emph{Advances in neural information processing systems}, 35:\penalty0 25278--25294, 2022.

\bibitem[Schulman et~al.(2015)Schulman, Levine, Abbeel, Jordan, and Moritz]{schulman2015trust}
John Schulman, Sergey Levine, Pieter Abbeel, Michael Jordan, and Philipp Moritz.
\newblock Trust region policy optimization.
\newblock In \emph{International conference on machine learning}, pages 1889--1897. PMLR, 2015.

\bibitem[Schulman et~al.(2017)Schulman, Wolski, Dhariwal, Radford, and Klimov]{schulman2017proximal}
John Schulman, Filip Wolski, Prafulla Dhariwal, Alec Radford, and Oleg Klimov.
\newblock Proximal policy optimization algorithms.
\newblock \emph{arXiv preprint arXiv:1707.06347}, 2017.

\bibitem[Scontras et~al.(2021)Scontras, Tessler, and Franke]{scontras2021practical}
Gregory Scontras, Michael~Henry Tessler, and Michael Franke.
\newblock A practical introduction to the rational speech act modeling framework.
\newblock \emph{arXiv preprint arXiv:2105.09867}, 2021.

\bibitem[Selvaraju et~al.(2016)Selvaraju, Das, Vedantam, Cogswell, Parikh, and Batra]{selvaraju2016grad}
Ramprasaath~R Selvaraju, Abhishek Das, Ramakrishna Vedantam, Michael Cogswell, Devi Parikh, and Dhruv Batra.
\newblock Grad-cam: Why did you say that?
\newblock \emph{arXiv preprint arXiv:1611.07450}, 2016.

\bibitem[Shen et~al.(2023)Shen, Jin, Huang, Liu, Dong, Guo, Wu, Liu, and Xiong]{shen2023large}
Tianhao Shen, Renren Jin, Yufei Huang, Chuang Liu, Weilong Dong, Zishan Guo, Xinwei Wu, Yan Liu, and Deyi Xiong.
\newblock Large language model alignment: A survey.
\newblock \emph{arXiv preprint arXiv:2309.15025}, 2023.

\bibitem[Shi et~al.(2024)Shi, Beltran~Velez, Nazaret, Zheng, Garriga-Alonso, Jesson, Makar, and Blei]{shi2024hypothesis}
Claudia Shi, Nicolas Beltran~Velez, Achille Nazaret, Carolina Zheng, Adri{\`a} Garriga-Alonso, Andrew Jesson, Maggie Makar, and David Blei.
\newblock Hypothesis testing the circuit hypothesis in llms.
\newblock \emph{Advances in Neural Information Processing Systems}, 37:\penalty0 94539--94567, 2024.

\bibitem[Sobel(2020)]{sobel2020signaling}
Joel Sobel.
\newblock Signaling games.
\newblock \emph{Complex social and behavioral systems: Game theory and agent-based models}, pages 251--268, 2020.

\bibitem[Szabadv{\'a}ry et~al.(2025)Szabadv{\'a}ry, L{\"o}fstr{\"o}m, Johansson, S{\"o}nstr{\"o}d, Ahlberg, and Carlsson]{szabadvary2025classification}
Johan~Hallberg Szabadv{\'a}ry, Tuwe L{\"o}fstr{\"o}m, Ulf Johansson, Cecilia S{\"o}nstr{\"o}d, Ernst Ahlberg, and Lars Carlsson.
\newblock Classification with reject option: Distribution-free error guarantees via conformal prediction.
\newblock \emph{Machine Learning with Applications}, page 100664, 2025.

\bibitem[Takmaz et~al.(2023)Takmaz, Brandizzi, Giulianelli, Pezzelle, and Fern{\'a}ndez]{takmaz2023speaking}
Ece Takmaz, Nicolo' Brandizzi, Mario Giulianelli, Sandro Pezzelle, and Raquel Fern{\'a}ndez.
\newblock Speaking the language of your listener: Audience-aware adaptation via plug-and-play theory of mind.
\newblock \emph{arXiv preprint arXiv:2305.19933}, 2023.

\bibitem[Tayebati et~al.(2025)Tayebati, Kumar, Darabi, Jayasuriya, Krishnan, and Trivedi]{tayebati2025learning}
Sina Tayebati, Divake Kumar, Nastaran Darabi, Dinithi Jayasuriya, Ranganath Krishnan, and Amit~Ranjan Trivedi.
\newblock Learning conformal abstention policies for adaptive risk management in large language and vision-language models.
\newblock \emph{arXiv preprint arXiv:2502.06884}, 2025.

\bibitem[Teneggi and Sulam(2024)]{teneggi2024testing}
Jacopo Teneggi and Jeremias Sulam.
\newblock Testing semantic importance via betting.
\newblock \emph{Advances in Neural Information Processing Systems}, 37:\penalty0 76450--76499, 2024.

\bibitem[Teneggi et~al.(2022{\natexlab{a}})Teneggi, Bharti, Romano, and Sulam]{teneggi2022shap}
Jacopo Teneggi, Beepul Bharti, Yaniv Romano, and Jeremias Sulam.
\newblock Shap-xrt: The shapley value meets conditional independence testing.
\newblock \emph{arXiv preprint arXiv:2207.07038}, 2022{\natexlab{a}}.

\bibitem[Teneggi et~al.(2022{\natexlab{b}})Teneggi, Luster, and Sulam]{teneggi2022fast}
Jacopo Teneggi, Alexandre Luster, and Jeremias Sulam.
\newblock Fast hierarchical games for image explanations.
\newblock \emph{IEEE Transactions on Pattern Analysis and Machine Intelligence}, 45\penalty0 (4):\penalty0 4494--4503, 2022{\natexlab{b}}.

\bibitem[Treviso and Martins(2020)]{treviso2020explanation}
Marcos~V Treviso and Andr{\'e}~FT Martins.
\newblock The explanation game: Towards prediction explainability through sparse communication.
\newblock \emph{arXiv preprint arXiv:2004.13876}, 2020.

\bibitem[Tsai and Carroll(2020)]{tsai2020logic}
Chun-Hua Tsai and John~M Carroll.
\newblock Logic and pragmatics in ai explanation.
\newblock In \emph{International Workshop on Extending Explainable AI Beyond Deep Models and Classifiers}, pages 387--396. Springer, 2020.

\bibitem[Tu et~al.(2025)Tu, Lin, Tian, Zhang, Li, Fu, Xu, He, Lan, Jiang, et~al.]{tu2025enhancing}
Songjun Tu, Jiahao Lin, Xiangyu Tian, Qichao Zhang, Linjing Li, Yuqian Fu, Nan Xu, Wei He, Xiangyuan Lan, Dongmei Jiang, et~al.
\newblock Enhancing llm reasoning with iterative dpo: A comprehensive empirical investigation.
\newblock \emph{arXiv preprint arXiv:2503.12854}, 2025.

\bibitem[Van~Fraassen(1980)]{van1980scientific}
Bas~C Van~Fraassen.
\newblock \emph{The scientific image}.
\newblock Oxford University Press, 1980.

\bibitem[Vaswani et~al.(2017)Vaswani, Shazeer, Parmar, Uszkoreit, Jones, Gomez, Kaiser, and Polosukhin]{vaswani2017attention}
Ashish Vaswani, Noam Shazeer, Niki Parmar, Jakob Uszkoreit, Llion Jones, Aidan~N Gomez, Lukasz Kaiser, and Illia Polosukhin.
\newblock Attention is all you need.
\newblock \emph{Advances in neural information processing systems}, 30, 2017.

\bibitem[Vilone and Longo(2021)]{vilone2021notions}
Giulia Vilone and Luca Longo.
\newblock Notions of explainability and evaluation approaches for explainable artificial intelligence.
\newblock \emph{Information Fusion}, 76:\penalty0 89--106, 2021.

\bibitem[Wah et~al.(2011)Wah, Branson, Welinder, Perona, and Belongie]{wah2011caltech}
Catherine Wah, Steve Branson, Peter Welinder, Pietro Perona, and Serge Belongie.
\newblock The caltech-ucsd birds-200-2011 dataset.
\newblock 2011.

\bibitem[Wang et~al.(2018)Wang, Chen, Yang, Wu, Wu, and Xie]{wang2018reinforcement}
Xiting Wang, Yiru Chen, Jie Yang, Le~Wu, Zhengtao Wu, and Xing Xie.
\newblock A reinforcement learning framework for explainable recommendation.
\newblock In \emph{2018 IEEE international conference on data mining (ICDM)}, pages 587--596. IEEE, 2018.

\bibitem[Wang et~al.(2024)Wang, Zhang, Guo, and Shen]{wang2024gradient}
Yongjie Wang, Tong Zhang, Xu~Guo, and Zhiqi Shen.
\newblock Gradient based feature attribution in explainable ai: A technical review.
\newblock \emph{arXiv preprint arXiv:2403.10415}, 2024.

\bibitem[Wanner et~al.(2024)Wanner, Van~Durme, and Dredze]{wanner2024dndscore}
Miriam Wanner, Benjamin Van~Durme, and Mark Dredze.
\newblock Dndscore: Decontextualization and decomposition for factuality verification in long-form text generation.
\newblock \emph{arXiv preprint arXiv:2412.13175}, 2024.

\bibitem[Wilson and Sperber(2012)]{wilson2012linguistic}
Deirdre Wilson and Dan Sperber.
\newblock Linguistic form and relevance.
\newblock \emph{Wilson \& Sperber (Eds.), Meaning and Relevance}, pages 149--168, 2012.

\bibitem[Wu et~al.(2024)Wu, Keoliya, Chen, Velingker, Li, Getzen, Long, Naik, Parikh, and Wong]{wu2024discret}
Yinjun Wu, Mayank Keoliya, Kan Chen, Neelay Velingker, Ziyang Li, Emily~J Getzen, Qi~Long, Mayur Naik, Ravi~B Parikh, and Eric Wong.
\newblock Discret: Synthesizing faithful explanations for treatment effect estimation.
\newblock \emph{Proceedings of machine learning research}, 235:\penalty0 53597, 2024.

\bibitem[Yu et~al.(2022)Yu, Wang, Vasudevan, Yeung, Seyedhosseini, and Wu]{yu2022coca}
Jiahui Yu, Zirui Wang, Vijay Vasudevan, Legg Yeung, Mojtaba Seyedhosseini, and Yonghui Wu.
\newblock Coca: Contrastive captioners are image-text foundation models.
\newblock \emph{arXiv preprint arXiv:2205.01917}, 2022.

\bibitem[Yuksekgonul et~al.(2022)Yuksekgonul, Wang, and Zou]{yuksekgonul2022post}
Mert Yuksekgonul, Maggie Wang, and James Zou.
\newblock Post-hoc concept bottleneck models.
\newblock \emph{arXiv preprint arXiv:2205.15480}, 2022.

\bibitem[Zermelo(1929)]{zermelo1929berechnung}
Ernst Zermelo.
\newblock Die berechnung der turnier-ergebnisse als ein maximumproblem der wahrscheinlichkeitsrechnung.
\newblock \emph{Mathematische Zeitschrift}, 29\penalty0 (1):\penalty0 436--460, 1929.

\bibitem[Zhang et~al.(2020)Zhang, Liao, and Bellamy]{zhang2020effect}
Yunfeng Zhang, Q~Vera Liao, and Rachel~KE Bellamy.
\newblock Effect of confidence and explanation on accuracy and trust calibration in ai-assisted decision making.
\newblock In \emph{Proceedings of the 2020 conference on fairness, accountability, and transparency}, pages 295--305, 2020.

\end{thebibliography}

\newpage
\clearpage
\appendix
\renewcommand\thefigure{\thesection.\arabic{figure}}
\renewcommand\thetable{\thesection.\arabic{table}}

\setcounter{figure}{0}
\section{\label{supp:experiments}Experimental Details}
In this section, we include further details on the experimental setup that were omitted from the main text for clarity of presentation. All experiments were run on a private server equipped with 8 NVIDIA RTX A5000 with 24GB of memory, and 500GB of RAM memory. 

\subsection{\label{supp:models}Model architectures}
In this section, we describe our speaker and listener model architectures.

\begin{figure}[h]
\centering
    \includegraphics[width=0.6\linewidth]{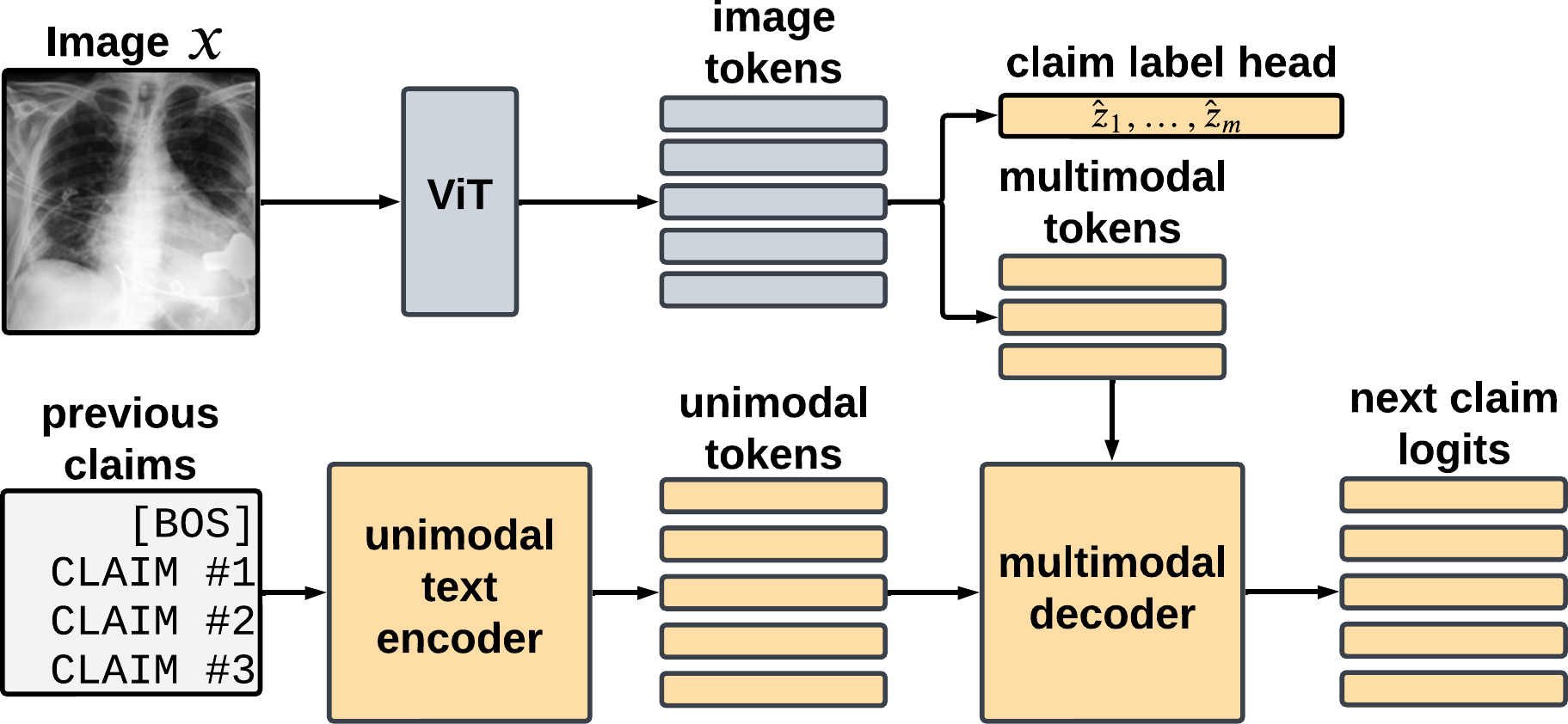}
    \caption{\label{fig:speaker_model}Illustration of the speaker model architecture. Gray elements are pretrained and frozen, yellow elements are learned.}
\end{figure}

\smallskip
\noindent\textbf{Speaker models}\\
\noindent We implement our speaker models starting from the CoCa \cite{yu2022coca} architecture, which consists of a unimodal text encoder followed by a multimodal decoder with cross-attention. Recall that utterances are composed of labeled claims, i.e. $u = [(c_1,\hat{z}_1), \dots, (c_l,\hat{z}_l)] \in (\C \times \{-1,+1\})^{\leq l}$, where $\hat{z}_j$ indicates if a claim is true or false for input $x$. Then, we include an attention head to predict these as illustrated in \cref{fig:speaker_model}. In particular, note that the signs are independent of the previous claims in the utterance, given the image tokens from the pretrained ViT encoder. That is, we parametrize the likelihood of the speaker model as
\begin{equation}
    P_S(c_j,\hat{z}_j \mid c_{<j}, h(x)) = P_S(c_j \mid c_{<j}, h(x))~P_S(\hat{z}_j \mid h(x)),
\end{equation}
where $c_{<j} = (c_1, \dots, c_{j-1})$ are the preceding claims in utterance $u$. Finally, note that the number and width of the multimodal tokens need not be the same as the image tokens. In this work, we use the following architecture parameters across all experiments: 6 unimodal and multimodal layers with a token width of 256, 4 attention layers and 4 attention pooling heads, and a number of multimodal tokens equal to half the maximum utterance length. We remark that the claim label head is not trained with strong supervision, but with preference data only.

\smallskip
\noindent\textbf{Listener models}\\
\noindent Listener models are bidirectional transformers \cite{devlin2019bert} with 12 layers, token width of 256, and 4 attention layers.

\subsection{\label{supp:hyperparameters}Behavior of \cref{algo:procedure} as a Function of Hyperparameters}
\begin{figure}[t]
\centering
\includegraphics[width=\linewidth]{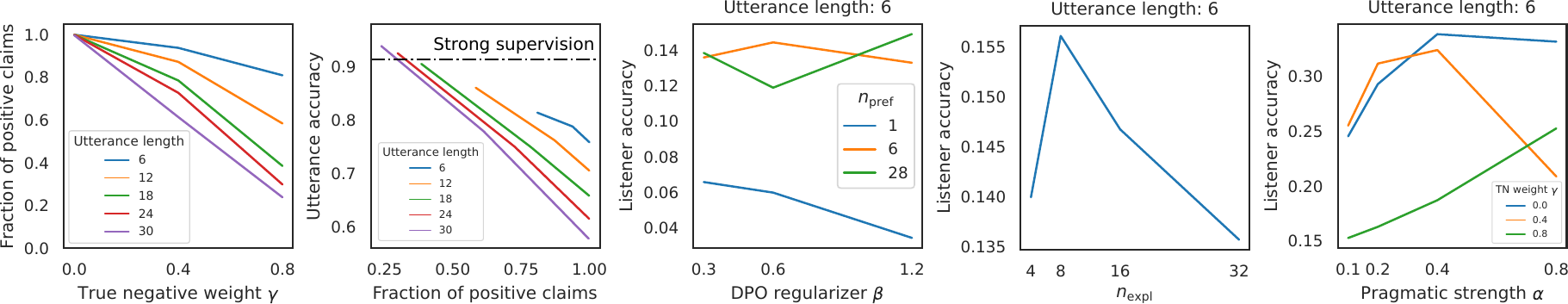}
\caption{\label{fig:cub_hyperparams}Summary of hyperparameter sweeps for our training procedure on the CUB dataset.}
\end{figure}

In this section, we discuss the behavior of our training procedure as a function of hyperparameters, as summarized in \cref{fig:cub_hyperparams} for the CUB dataset.

\smallskip
\noindent\textbf{True negative weight $\gamma$:} This hyperparameter controls the weight of true negative claims in the fidelity score in \cref{eq:fidelity}.  As $\gamma \to 0$, utterances with negative claims are downweighted, favoring positive claims. Indeed, we can see that the fraction of positive claims decreases as $\gamma$ increases, across all utterance lengths.

\smallskip
\noindent\textbf{Accuracy of utterances:} Next, we investigate the effect of the fraction of positive claims on the accuracy of the utterances, i.e., the correctness of the predicted claim labels. We can see that accuracy decreases with the fraction of positive claims, which is due to the fact that the vocabulary of claims is usually very imbalanced. In the CUB dataset, for example, images have on average $\approx 30$ positive claims out of the 312 in the dataset. Furthermore, we include the accuracy of the claim label attention head when trained with strong supervision as a baseline value to verify that the pretrained embeddings from the fixed classifier contain the semantic information of the claims. We can see that strong supervision does lead to improved accuracy compared to weak supervision through preference pairs.

\smallskip
\noindent\textbf{DPO regularization parameter $\beta$:} This parameter controls the KL divergence between the updated speaker $S\at{t+1}$ and its reference $S\at{t}$. Intuitively, it prevents the speaker model from changing abruptly in one iteration. The third panel in \cref{fig:cub_hyperparams} shows listener accuracy as a function of $\beta$, stratified by the number of preferences $n_{\pref}$ used to update the speaker at each iteration. We can see that the optimal value of $\beta$ changes depending on $n_{\pref}$, where greater values of $n_{\pref}$ lead to larger values of $\beta$ that maximize the accuracy of the listener. We also remark that increasing $n_{\pref}$ increases the computational cost of each iteration, since it requires sampling more candidate utterances.

\smallskip
\noindent\textbf{Number of explanations $n_{\expl}$:} This is the number of utterances sampled per input when updating the listener $L\at{t}$. In this case, we fix $\gamma=0.4$, $n_{\pref}=6$, $\beta=0.6$ and vary $n_{\expl}$ only for utterances of length 6. We can see that the accuracy of the listener peaks at $n_{\expl}$ and then decreases for larger values. This finding suggests that sampling too many utterances to construct $D_{\expl}\at{t}$ may hurt optimization. 

\smallskip
\noindent\textbf{Pragmatic strength $\alpha$:} Finally, we fix all remaining parameters and study listener accuracy as a function of $\alpha$, i.e., the weight of the listener's choice in ranking pairs of utterances, and $\gamma$. We see that for small values of $\gamma$, the accuracy of the listener has a concave shape, whereas for large values of $\gamma$, the hump does not appear for the values of $\alpha$ we considered. As is generally true with regularizers, our findings confirm that there exists a tradeoff between fidelity and utility, and overweighting utility may be detrimental to the accuracy of the listener model.

\setcounter{figure}{0}
\section{Additional results}
In this section, we include additional results omitted from the main text for the sake of conciseness of presentation.

\subsection{\label{supp:accuracy}Accuracy of listener models}
\begin{figure}[h]
\centering
\subcaptionbox{CUB.}{\includegraphics[height=0.14\linewidth]{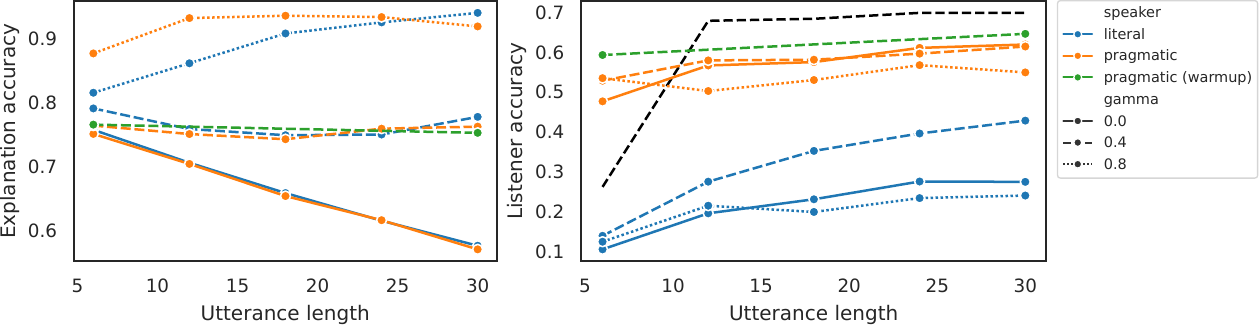}}
\hfill
\subcaptionbox{CheXpert.}{\includegraphics[height=0.14\linewidth]{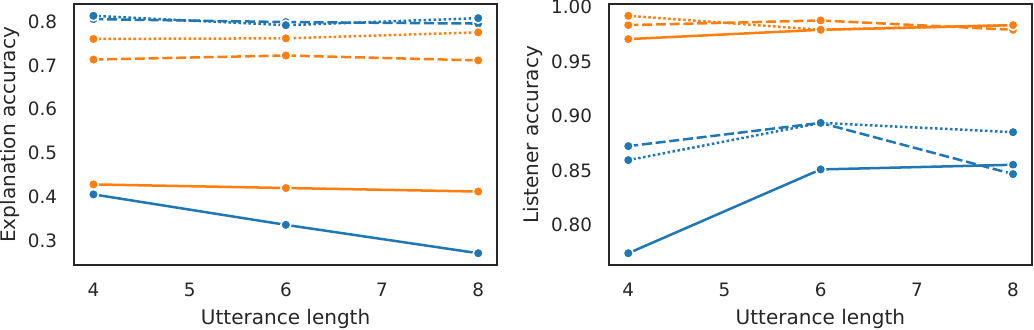}}
\caption{\label{fig:supp_accuracy}Explanation and listener accuracy as a function of utterance length and true negative weight $\gamma$ for the CUB and CheXpert datasets.}
\end{figure}

Here, we expand on the results presented in \cref{table:accuracy} for the CUB and CheXpert datasets. \cref{fig:supp_accuracy} illustrates explanation accuracy (i.e., the accuracy of the predicted binary labels $\z \in \{-1,1\}$) and listener accuracy as a function of utterance length and true negative weight $\gamma$.

For both datasets, we can appreciate that the listener accuracy with a pragmatic speaker is consistently higher than with a literal listener across all utterance lengths. As discussed in the main text, V-IP outperforms pragmatic listeners on CUB for utterances longer than 12 claims. This is consistent with the intuition that V-IP solves the reward optimization problem in \cref{eq:objective} with a listener that can be differentiated through the speaker, which is a stronger condition than our setting. We also remark that V-IP is a two-step procedure that first trains a concept classifier and subsequently builds interpretable predictors, and that the first step requires concept annotations for supervised training. For explanation accuracy, we observe that $(i)$ increasing $\gamma$ allows for more negative claims to be included in the utterances, which generally increase accuracy, and $(ii)$ in the CUB dataset, there is little difference in accuracy between the literal and pragmatic speakers across all utterance lengths, whereas literal speakers outperform pragmatic ones in the CheXpert dataset. This trend highlights the possible tradeoff between fidelity and utility in the reward.

\subsection{\label{supp:diversity}Diversity of utterances}
In this section, we study the diversity of the utterances generated by a literal and a pragmatic speaker on the CUB dataset. We compute the similarity of positive and negative claims between pairs of utterances via intersection over union. That is, two utterances are similar if they contain the same positive and negative claims.

\begin{figure}[h]
    \centering
    \includegraphics[width=\linewidth]{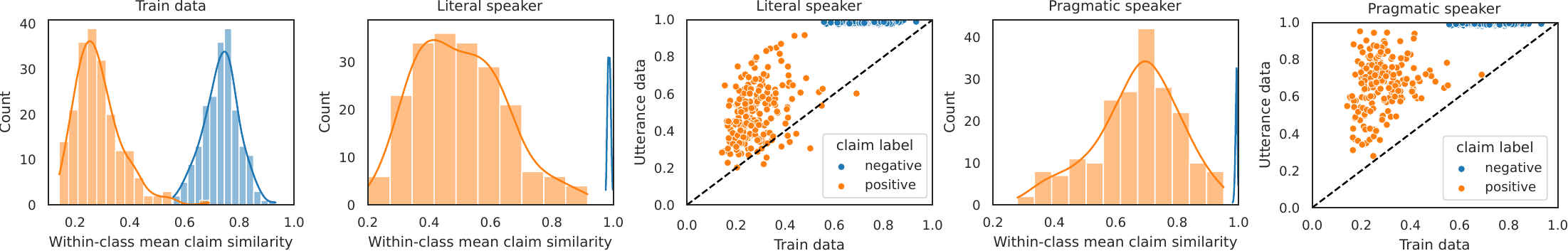}
    \caption{\label{fig:supp_diversity}Distribution of within-class claim similarity of utterances composed of 6 claims, generated by a literal and a pragmatic speaker on the CUB dataset.}
\end{figure}

\cref{fig:supp_diversity} compares the distribution of the mean within-class similarity of the ground-truth annotations with the generated utterances. We can see that in the training data, negative labels are generally more similar than positive ones. Furthermore, both speakers increase the similarity of all claims, both positive and negative, especially negative ones, which appear to be mostly constant for all images with the same ground-truth label. In particular, the pragmatic speaker increases the average claim similarity of all classes compared to the ground-truth data. This is not surprising, as the speaker needs to cooperate with the listener to construct utterances that convey the same message, and this may be achieved by finding constant explanations for each label.

\subsection{\label{supp:user_study_results}User study}
\begin{wrapfigure}{R}{0.3\linewidth}
    \vspace{-20pt}
    \centering
    \includegraphics[width=\linewidth]{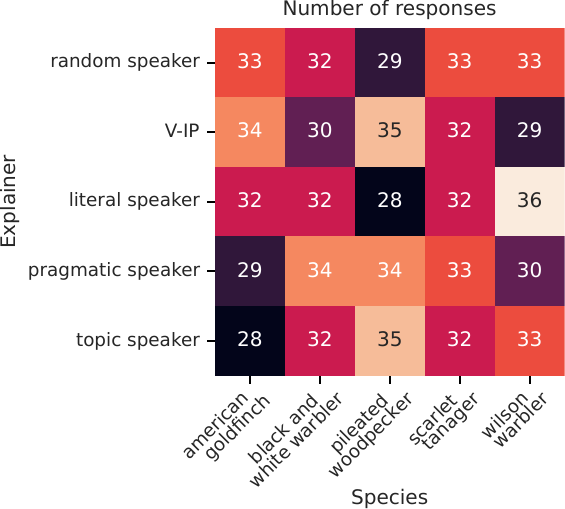}
    \caption{\label{fig:supp_user_study_responses}Number of registered responses per explainer-species combination.}
\end{wrapfigure}

In this section, we expand on the discussion of the results of our user study. First, \cref{fig:supp_user_study_responses} includes the distribution of responses collected during the study. Note that, without a centralized server to keep track of each participant's assignments, trials are not exactly balanced because each participant sees a random subset of 4 out of 5 species for each condition. For 40 participants (the size of our cohort) the expected number of responses per explainer-species pair is 32, and we can see this is the case from \cref{fig:supp_user_study_responses}.

Next, recall that, by design, we select 5 bird species that share some key traits in order to make the classification task nontrivial. We investigate whether the variability in classification performance across different species may be explained by how distinctive the claims included in the utterances by each explainer are. For example, consider the species American goldfinch and Wilson warbler, which are both yellow birds (see \cref{fig:supp_user_study_species}). Visually comparing the two, one can see that the most distinctive traits of the American goldfinch are black wings, a black forehead, and an orange bill. The utterances generated by the literal speaker (6\% accuracy), the pragmatic speaker (52\% accuracy), and the topic speaker (79\% accuracy) are:

\begin{center}
\begin{tabular}{ccc}
    Literal speaker (6\%)           & Pragmatic speaker (52\%)          & Topic speaker (79\%)\\
    \cmidrule(r){1-1}               \cmidrule(r){2-2}                   \cmidrule(r){3-3}
    bill shorter than head (yes)    & black eye (yes)                   & {\color{green}yellow belly (yes)}\\
    solid belly (yes)               & {\color{green}yellow belly (yes)} & {\color{green}is yellow (yes)}\\
    solid breast (yes)              & {\color{green}is yellow (yes)}    & {\color{green}solid belly (yes)}\\
    black eye (yes)                 & yellow underparts (yes)           & {\color{orange}\textbf{black wing (yes)}}\\
    cone bill (yes)                 & bill shorter than head (yes)      & solid breast (yes)\\
    perching-like (yes)             & {\color{green}solid belly (yes)}  & {\color{orange}\textbf{black forehead (yes)}}
\end{tabular}
\end{center}

Green claims are shared between the pragmatic and topic speakers, and bold orange claims are unique to the topic speaker. First, we can see that although the literal speaker includes correct claims about the appearance of the bird, they are not very informative (no claims mention yellow coloration). The pragmatic speaker does include claims about the coloration of the bird, but they can be mistaken for a Wilson warbler. We remark that the pragmatic listener model achieves a classification accuracy of 100\% for the American goldfinch, which is higher than the observed human accuracy of 52\%. This example captures the importance of aligning explanations to human preferences because the claims chosen by a pragmatic speaker may be difficult to understand for a human listener. We can see that the topic speaker does in fact mention both black wings and black forehead, allowing users to correctly identify the bird species.

Interestingly, the situation changes for the classification of Wilson warbler, where the pragmatic speaker achieves higher accuracy (60\%) compared to the literal speaker (44\%) and the topic speaker (36\%). The respective utterances are:

\begin{center}
\begin{tabular}{ccc}
    Literal speaker (44\%)                  & Pragmatic speaker (60\%)                              & Topic speaker (36\%)\\
    \cmidrule(r){1-1}                       \cmidrule(r){2-2}                                       \cmidrule(r){3-3}
    {\color{green}yellow underparts (yes)}  & {\color{orange}\textbf{all-purpose bill (yes)}}       & yellow throat (yes)\\
    {\color{green}yellow belly (yes)}       & {\color{green}yellow underparts (yes)}                & solid breast (yes)\\
    {\color{green}yellow breast (yes)}      & {\color{green}yellow belly (yes)}                     & solid belly (yes)\\
    is yellow (yes)                         & {\color{orange}\textbf{bill shorter than head (yes)}} & {\color{green}yellow breast (yes)}\\
    black eye (yes)                         & {\color{green}yellow breast (yes)}                    & {\color{green}yellow underparts (yes)}\\
    solid breast (yes)                      & black eye (yes)                                       & {\color{green}yellow belly (yes)}
\end{tabular}
\end{center}

Green claims are shared between all explainers, and orange claims are unique to the pragmatic speaker. In this case, we can see that features about the bill are useful for users to correctly identify the species of the bird, but these claims are excluded by the topic speaker because of the prior on head features, coloration, and patterns only. This stresses the importance of collecting human data to model their preferences.

Lastly, we compare utterances for the black and white warbler, which shares key features with the pileated woodpecker. For this bird species, the topic speaker achieves the highest accuracy (94\%), and V-IP beats the pragmatic speaker (80\% vs 71\%):

\begin{center}
\begin{tabular}{ccc}
    V-IP (80\%)             & Pragmatic speaker (71\%)                  & Topic speaker (94\%)\\
    \cmidrule(r){1-1}       \cmidrule(r){2-2}                           \cmidrule(r){3-3}
    Black wing (yes)        & Black eye (yes)                           & White nape (yes)\\
    All-purpose bill (yes)  & Bill about the same length as head (yes)  & White breast (yes)\\
    White belly (yes)       & All-purpose bill (yes)                    & Black eye (yes)\\
    Black leg (yes)         & {\color{red}White throat (yes)}                        & White throat (yes)\\
    Solid breast (yes)      & {\color{red}White underparts (yes)}                    & Brown upperparts (no)\\
    Notched tail (yes)      & Is white (yes)                            & {\color{orange}\textbf{Striped wing (yes)}}
\end{tabular}
\end{center}

\begin{figure}[h]
    \centering
    \includegraphics[width=\linewidth]{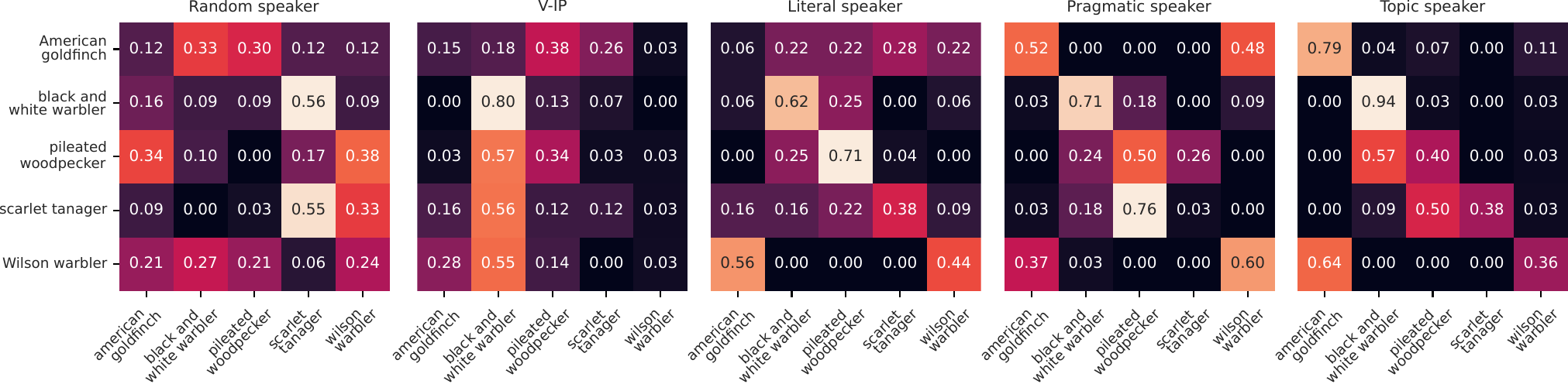}
    \caption{\label{fig:supp_user_study_confusion}Confusion matrices for all explainers and bird species.}
\end{figure}

In this case, utterances are more diverse across explainers, with different claims about the coloration (black and/or white) and features of the bill, head, and tail. First, we note that users misinterpret the utterance of the pragmatic speaker as referring to a pileated woodpecker (18\% misclassification rate), which does not have white underparts. We hypothesize this may be because the pileated woodpecker has a distinctive white throat, which is mentioned in the utterance of the pragmatic speaker. Second, we note that the topic speaker is the only one to mention the striped wings of the bird, which almost always disambiguate the choice. For the sake of completeness, \cref{fig:supp_user_study_confusion} includes the confusion matrices for all bird species and explainers.

\subsection{\label{supp:complexity}Computational complexity}
In this section, we include the runtimes of our algorithm as a function of utterance length across all datasets. We remark that our procedure involves sampling from the speaker at each iteration, first to construct preference pairs, and subsequently to sample explanations. We implement our speaker model as an autoregressive encoder-decoder architecture with multimodal cross-attention, which is known to have quadratic complexity in the sequence length \cite{vaswani2017attention}.

\begin{table}[h]
\caption{\label{table:complexity}Runtimes of our training procedure across all datasets. All runtimes are for one NVIDIA RTX A5000 with 24GB of memory. Results are reported as min/max ranges, where variability comes from the overall load of the server at the time of training. We exclude runtimes for CheXpert with 6 claims because the server was particularly overloaded, and runtimes are not trustworthy.}
\centering
\begin{minipage}{0.48\linewidth}
\small
\begin{tabular}{lcc}
    \toprule
    Dataset & Utterance length  & Runtime (hours)\\
    \midrule
    CUB     & 6                 & 4~/~8\\
    CUB     & 12                & 6~/~8\\
    CUB     & 18                & 7~/~9\\
    CUB     & 24                & 8~/~11\\
    CUB     & 30                & 11~/~16\\
    \bottomrule
\end{tabular}
\end{minipage}
\hfill
\begin{minipage}{0.48\linewidth}
\small
\begin{tabular}{lcc}
    \toprule
    Dataset         & Utterance length  & Runtime (hours)\\
    \midrule
    CheXpert        & 4                 & 4~/~9\\
    CheXpert$^*$    & 6                 & -\\
    CheXpert        & 8                 & 5~/~12\\
    CheXpert        & 12                & 6~/~14\\
    ImageNet        & 12                & 21~/~31\\
    \bottomrule
\end{tabular}
\end{minipage}
\end{table}

\setcounter{figure}{0}
\setcounter{table}{0}
\section{User study details}
In this section, we expand on details of the user study that were omitted from the main text for the sake of presentation. Throughout this section, ``pragmatic topic listener'' refers to a pragmatic listener with uniform prior topic preference on claims about the head, coloration, and patterns of the bird.

\subsection{\label{supp:user_study_flowchart}Flowchart}
\begin{figure}[h]
    \centering
    \includegraphics[width=\linewidth]{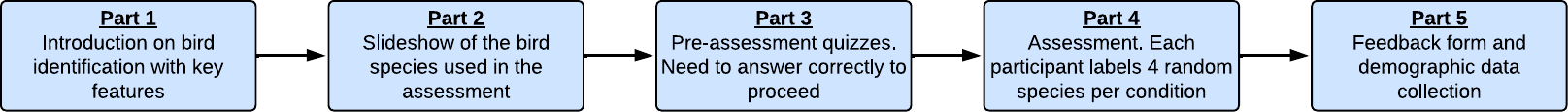}
    \caption{\label{fig:supp_flowchart}Flowchart of the user study.}
\end{figure}

We use jsPsych \cite{de2015jspsych} to implement the web browser interface of our user study. \cref{fig:supp_flowchart} illustrates the 4 parts of our survey. We remark that, in order to proceed to the assessment, participants need to answer the quizzes correctly. This is to ensure all participants have developed a basic understanding of the task and key features of the 5 bird species. We briefly describe each part of the survey with screenshots.

\subsubsection{Part 1: bird identification with key features}
\begin{figure}[h]
    \centering
    \includegraphics[width=\linewidth]{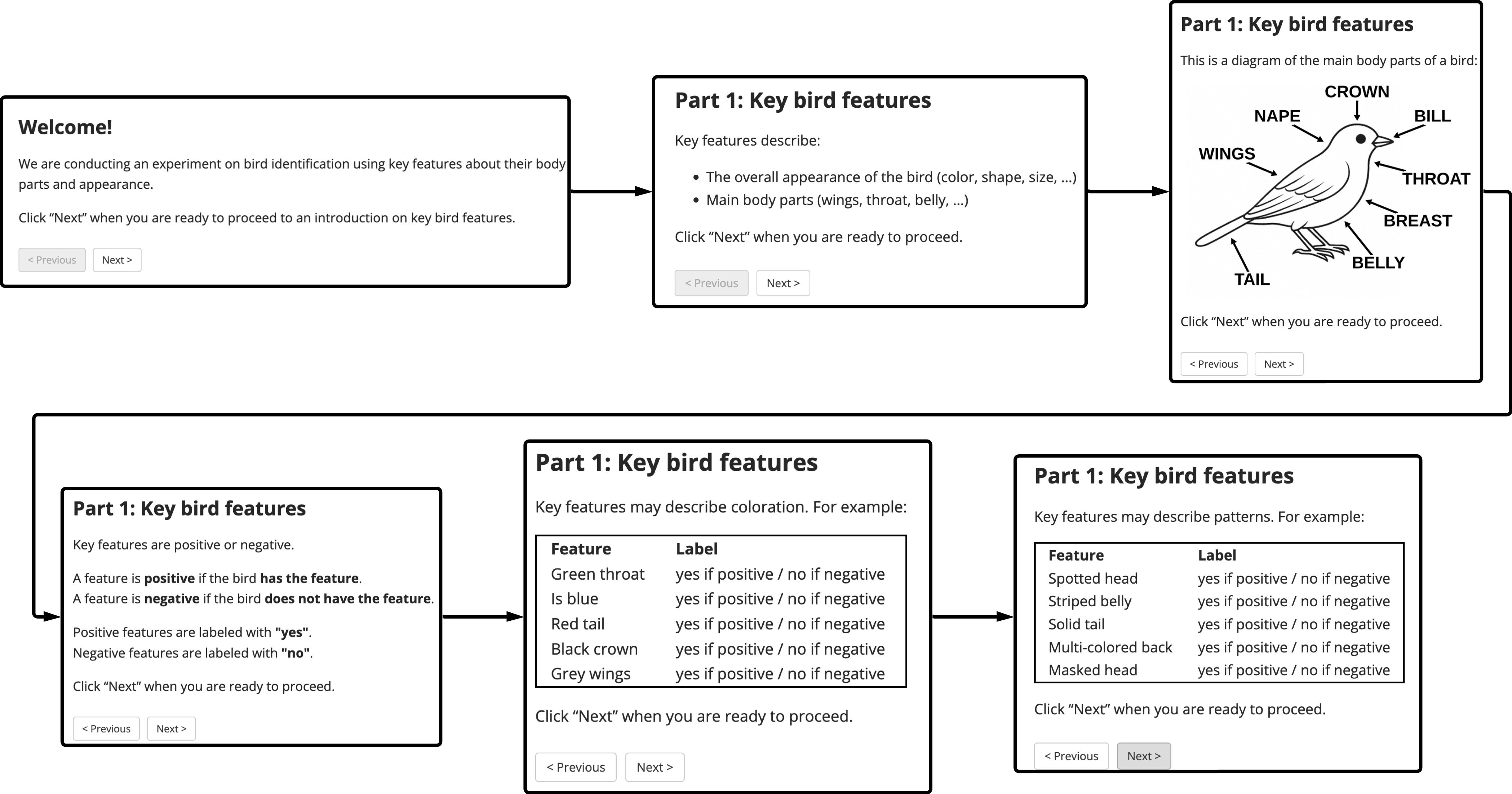}
    \caption{\label{fig:supp_user_study_part1}Screenshots of part 1 of the user study.}
\end{figure}

Participants are first introduced to the task of bird identification with key features. In the study, we use the same features as the ground-truth annotations in the CUB dataset. Participants are shown an illustration of the main body parts of a bird, and a few example features (see \cref{fig:supp_user_study_part1}). 

\subsubsection{Part 2: slideshow of bird species}
\begin{figure}[h]
    \centering
    \includegraphics[width=\linewidth]{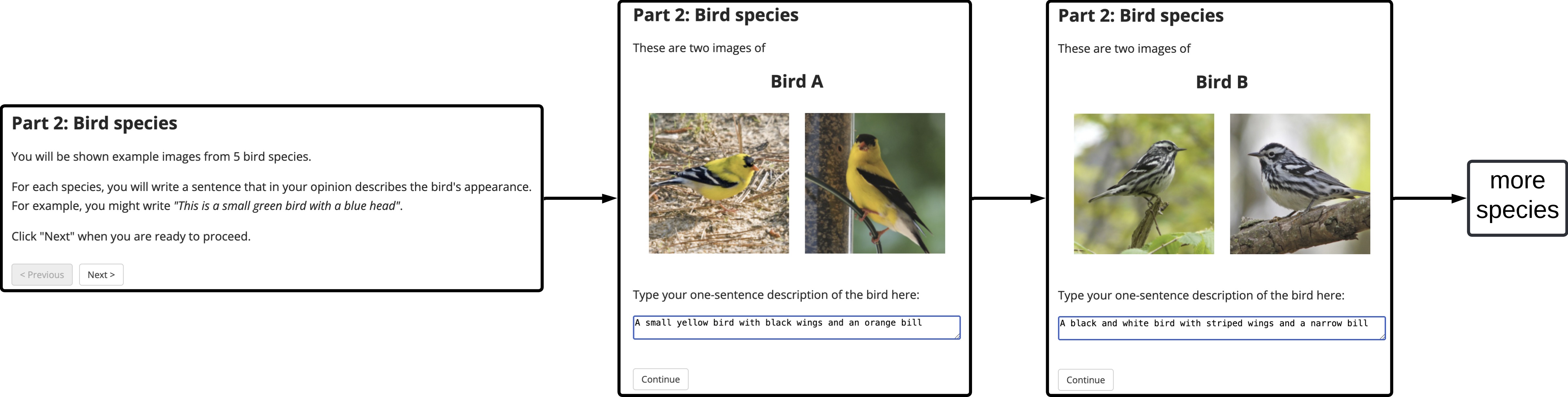}
    \caption{\label{fig:supp_user_study_part2}Screenshots of part 2 of the user study.}
\end{figure}

In part 2, participants are shown two example images for each bird species, and they are asked to summarize their appearance in one sentence. We remark that, in the study, bird species are referred to with the pseudonyms \verb|"Bird <<A-E>>"| instead of their names. This choice is motivated by two observations. First, the names of the bird species may reveal their most distinctive traits. For example, it is immediate to realize that a \emph{black and white warbler} should be black and white. Using uninformative identifiers removes bias in the species names. Second, names of bird species may be unfamiliar and confusing to participants without prior birdwatching experience (e.g., people may not know what a \emph{tanager} is). Our pseudonyms also remove this potential source of bias. \cref{fig:supp_user_study_part2} includes screenshots of this part of the survey.

\subsubsection{Part 3: pre-assessment quizzes}
\begin{figure}[h]
    \centering
    \includegraphics[width=\linewidth]{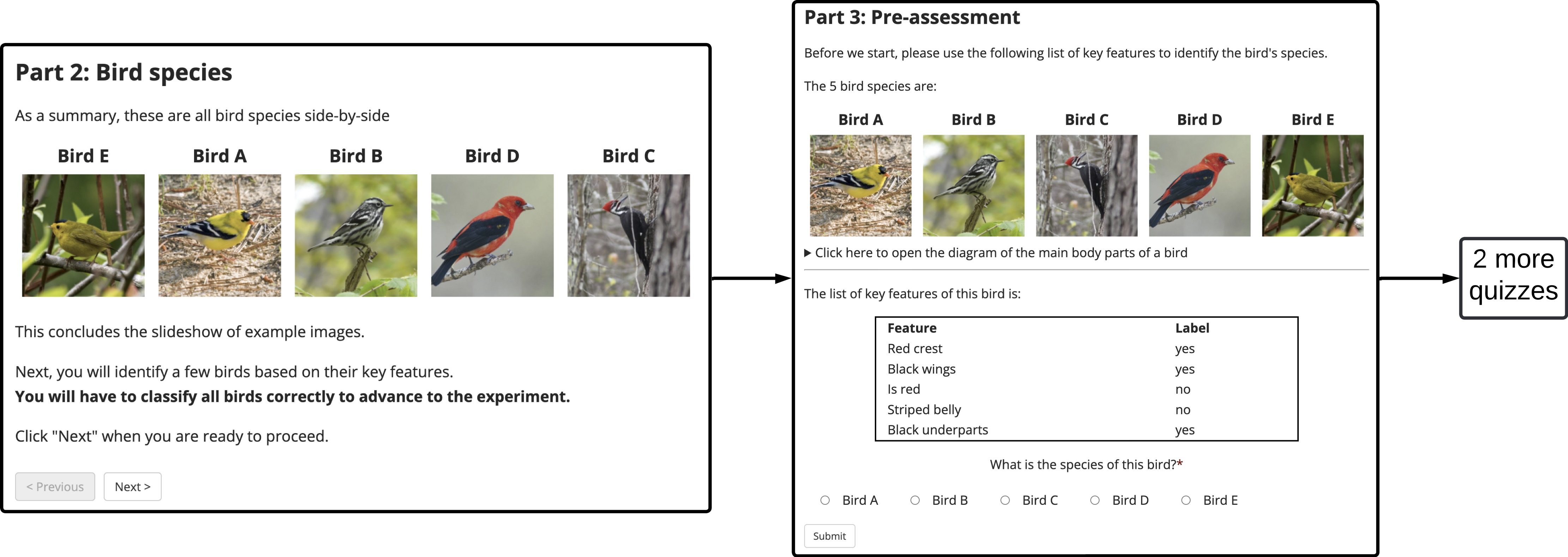}
    \caption{\label{fig:supp_user_study_part3}Screenshots of part 3 of the user study.}
\end{figure}

Before the assessment, participants need to solve three quizzes in part 3 of the survey. This is to familiarize themselves with the assessment interface, and to make sure they have gained a basic understanding of the task and the appearance of the 5 bird species. The list of key features used in the quizzes did not come from any of the explainers being tested, but they were designed by us, in order not to introduce any bias. We note that participants have access to a visual manual of the 5 bird species, and they can access the illustration of the main body parts of a bird at any time (see \cref{fig:supp_user_study_part3} for screenshots of part 3).

\subsubsection{Part 4: assessment}
\begin{figure}[h]
    \centering
    \includegraphics[width=\linewidth]{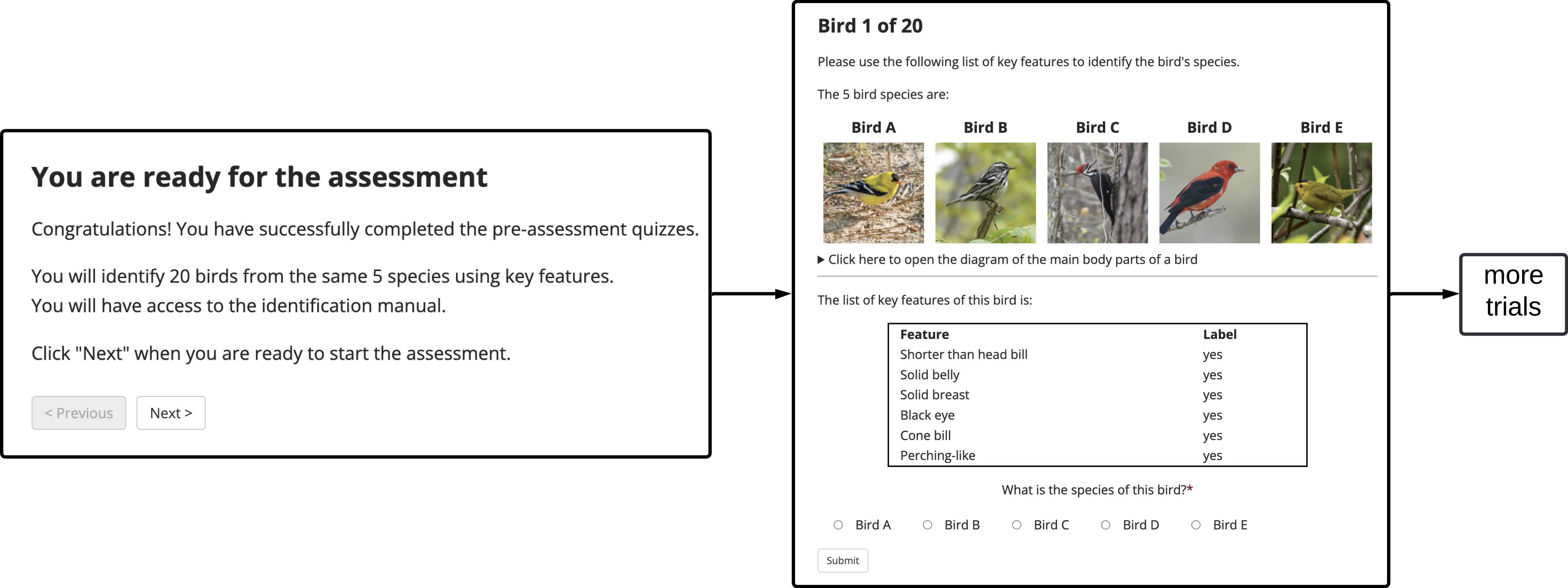}
    \caption{\label{fig:supp_user_study_part4}Screenshots of part 4 of the user study.}
\end{figure}

During the assessment, participants are asked to categorize 20 birds by only looking at the utterances generated by the explainers. The user interface is the same as for the quizzes, and we include screenshots in \cref{fig:supp_user_study_part4}.

\subsubsection{Part 5: demographics and feedback}
Lastly, participants are asked to share age group, gender, and prior birdwatching experience on a qualitative scale from no prior experience to professional birdwatcher. Participants are also encouraged to share feedback to help us improve future studies, which concludes the survey.

\subsection{\label{supp:user_study_species}Chosen bird species}
\begin{figure}[t]
\begin{minipage}{0.45\linewidth}
\includegraphics[width=0.9\linewidth]{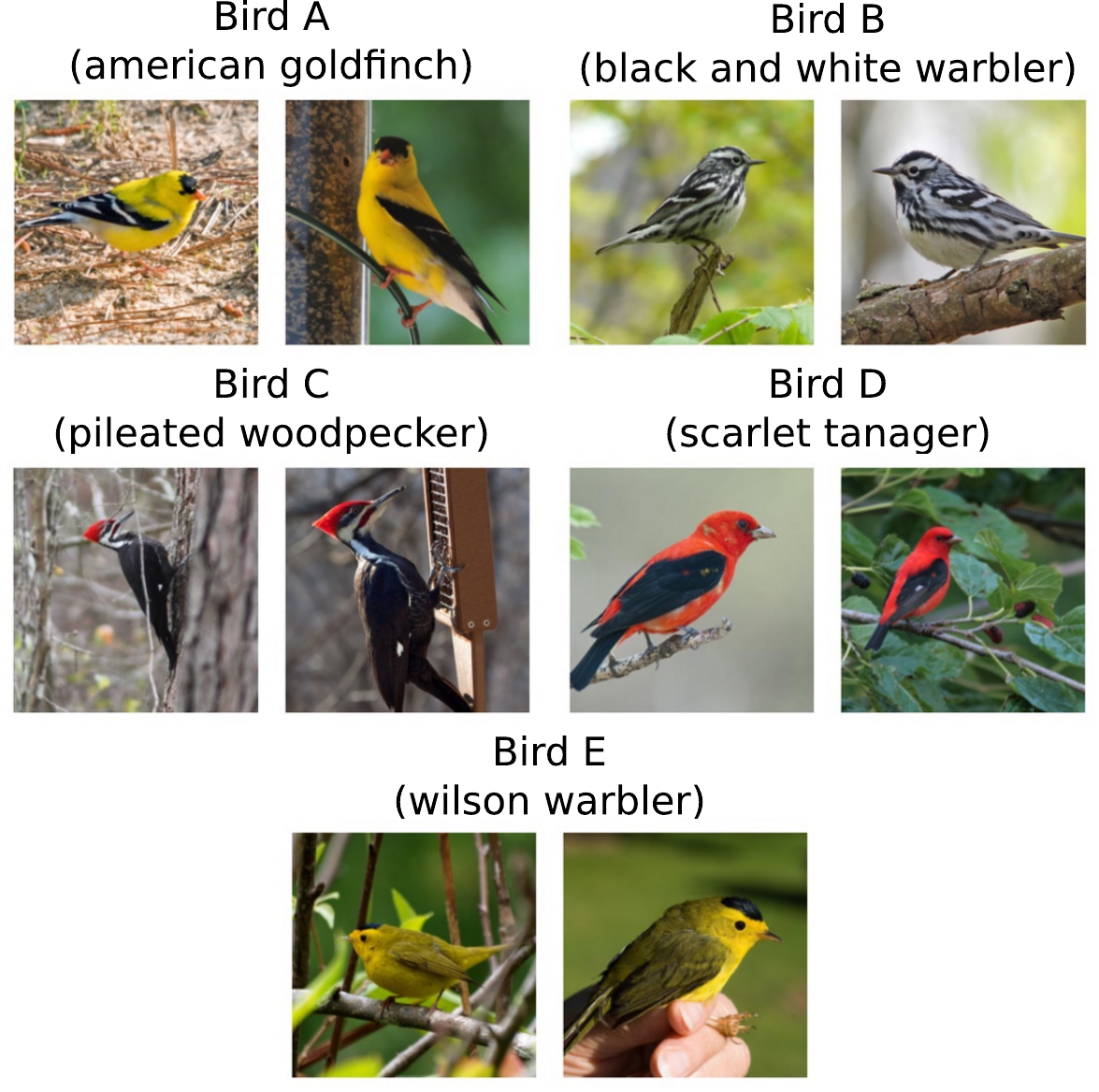}
\captionof{figure}{\label{fig:supp_user_study_species}Example images of the 5 bird species used in the user study. Note that species are referred to with pseudonyms ``Bird A'', ``Bird B'', ..., ``Bird E''.}
\end{minipage}
\hfill
\begin{minipage}{0.50\linewidth}
\captionof{table}{\label{table:supp_species}Classification accuracy of a pragmatic listener with no topic prior and a pragmatic listener with uniform topic prior on claims about the head features, coloration, and patterns of a bird.}
\small
\resizebox{\linewidth}{!}{%
\begin{tabular}{lcc}
    \toprule
                    & \multicolumn{2}{c}{Accuracy}\\
    \cmidrule(r){2-3}
    Bird species            & Pragmatic listener    & Pragmatic topic listener\\
    \midrule
    Pileated woodpecker     & 100\%                 & 100\%\\
    American goldfinch      & 100\%                 & 96\%\\
    Scarlet tanager         & 92\%                  & 100\%\\
    Black and white warbler & 100\%                 & 87\%\\
    Wilson warbler          & 83\%                  & 96\%\\
    \bottomrule
\end{tabular}}
\end{minipage}
\end{figure}

\cref{fig:supp_user_study_species} includes the two example images shown to participants of the study, and \cref{table:supp_species} reports the classification accuracy of the pragmatic listeners used to select them. Note that the chosen bird species have intersecting key features. For example, both Bird A and E are yellow, both Bird A, D, and E have black wings, and both Bird D and C are black and white. At the same time, all species have some unique traits: Bird A has an orange bill and black forehead, Bird B has striped wings, Bird C has a red crest, Bird D is red, and Bird E has a dark crown.

\subsection{\label{supp:user_study_cohort}Cohort details}
We recruited participants through Prolific (\url{https://prolific.com}) with the following inclusion criteria: any country, age range 18--65, primary language English, fluent in English, and without colorblindness. Participants were compensated \$12.00/hr prorated by completion time (average reward after completion was \$13.97/hr).

\begin{figure}[h]
    \centering
    \includegraphics[width=0.6\linewidth]{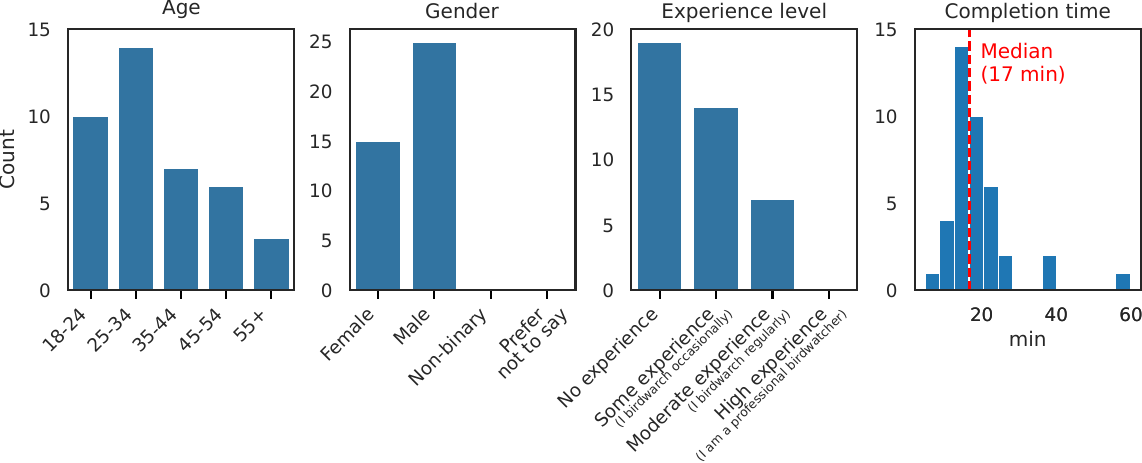}
    \caption{\label{fig:supp_user_study_cohor}Summary of demographics and completion times of our cohort of 40 participants.}
\end{figure}

\newpage
\clearpage

\setcounter{figure}{0}
\section{\label{supp:figures}Supplementary figures}
In this section, we include additional figures that were omitted from the main text for conciseness. Since listener models are bidirectional transformers, we include barplots that show the average attention weight of each claim in the utterance across all layers and attention heads.

\begin{figure}[h]
    \centering
    {\includegraphics[width=0.65\linewidth]{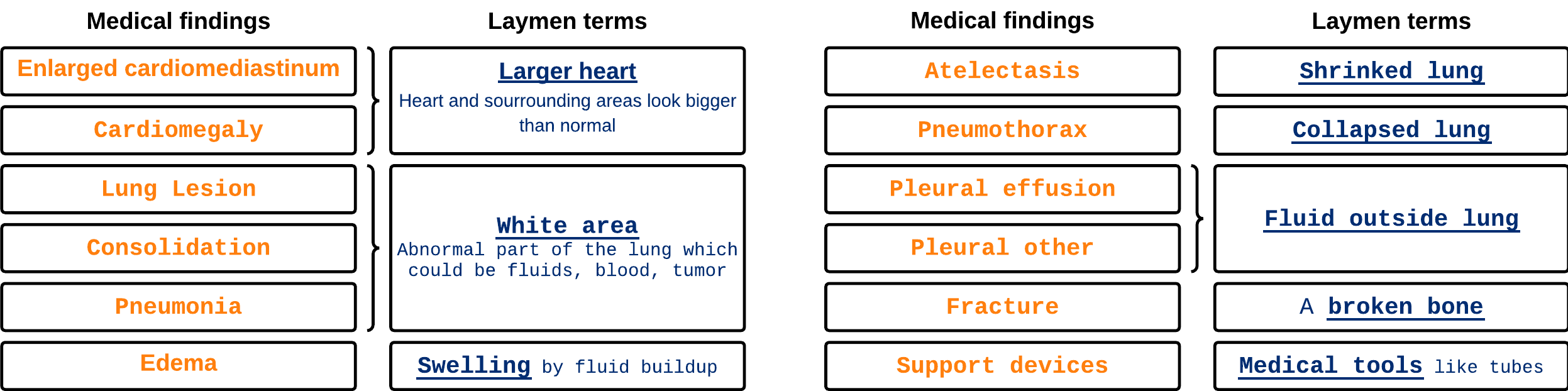}}
    \caption{\label{fig:chexpert_vocabulary}Mapping of medical claims in the CheXpert dataset to laymen friendly descriptions. The underlined terms are the abbreviations used in the figures in the main text of the manuscript.}
\end{figure}

\begin{figure}[h]
    \vspace{-20pt}
    \centering
    \includegraphics[width=0.65\linewidth]{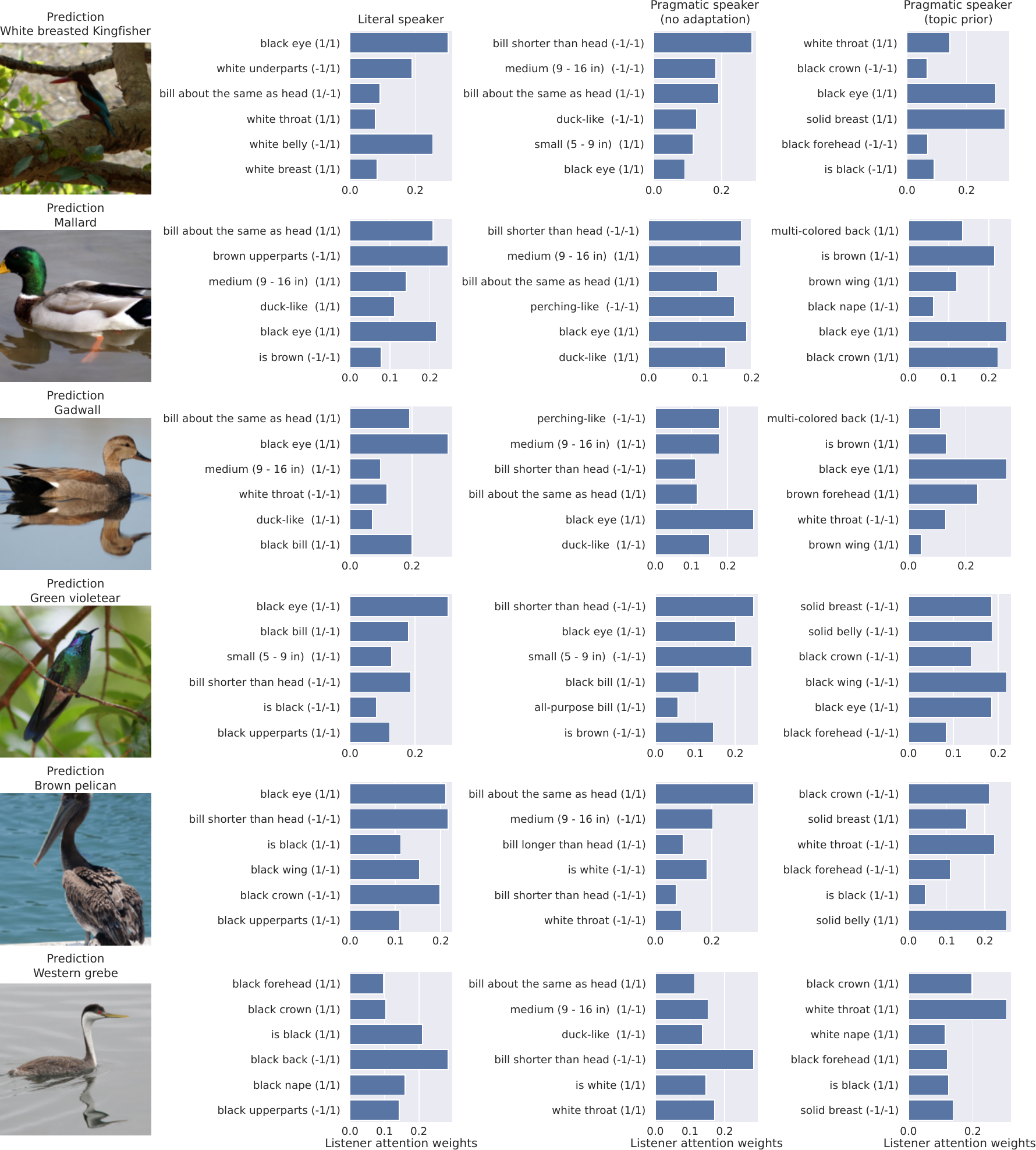}
    \caption{Additional results on the CUB dataset for a literal speaker, a pragmatic speaker with no adaptation, and a pragmatic speaker adapted to the topic prior described in the main text of the manuscript. We include cases where the classifier and all listeners correctly retrieve the ground-truth label of the image.}
\end{figure}

\begin{figure}[h]
    \vspace{-20pt}
    \centering
    \includegraphics[width=\linewidth]{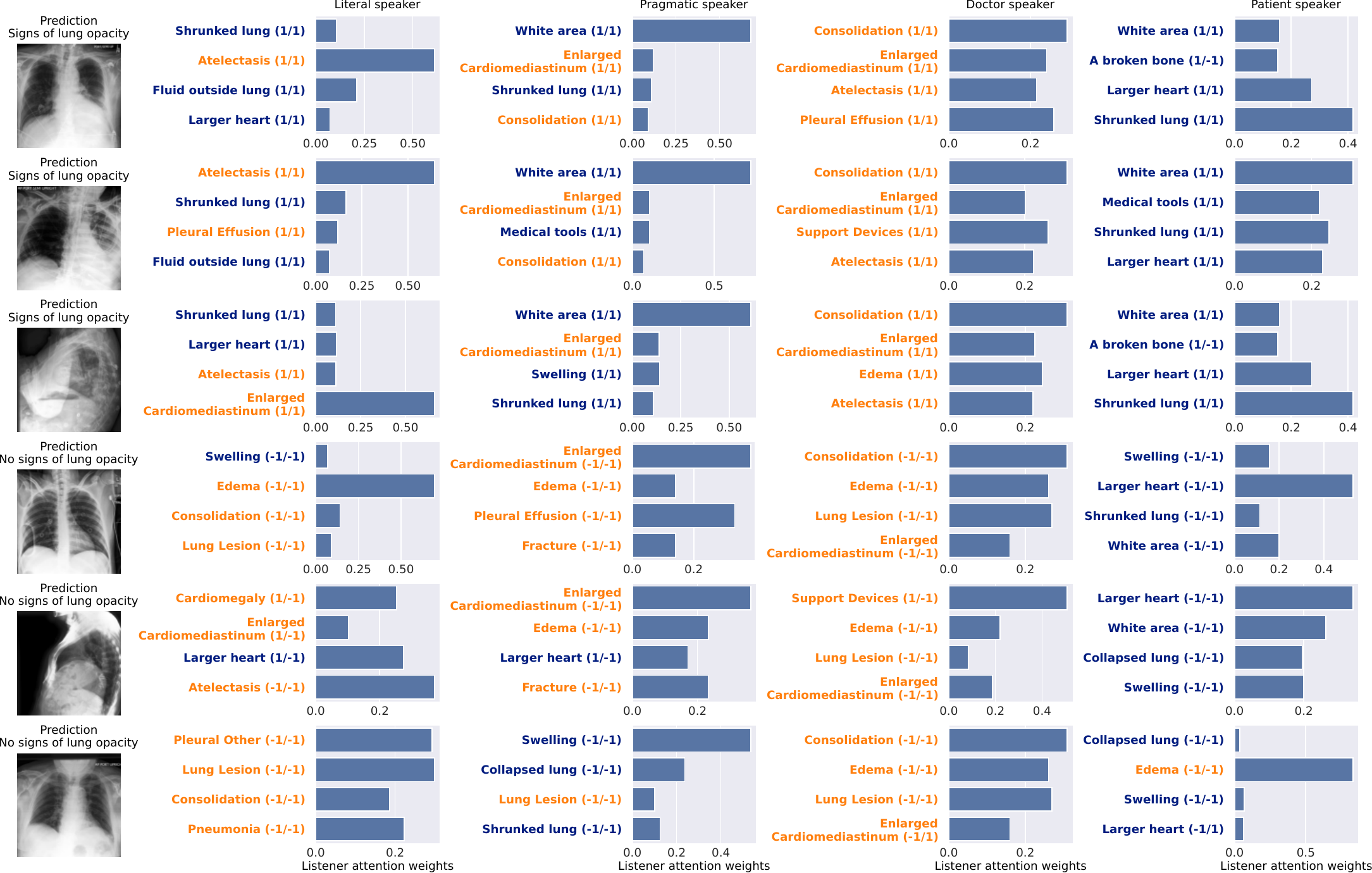}
    \caption{Additional results on the CheXpert dataset for a literal speaker, an unadapted pragmatic speaker, a doctor-preferred speaker, and a patient-preferred speaker. The blue and orange colors denote medical terms and layman-friendly descriptions, respectively.}
    \vspace{-10pt}
\end{figure}

\begin{figure}[h]
    \centering
    \includegraphics[width=\linewidth]{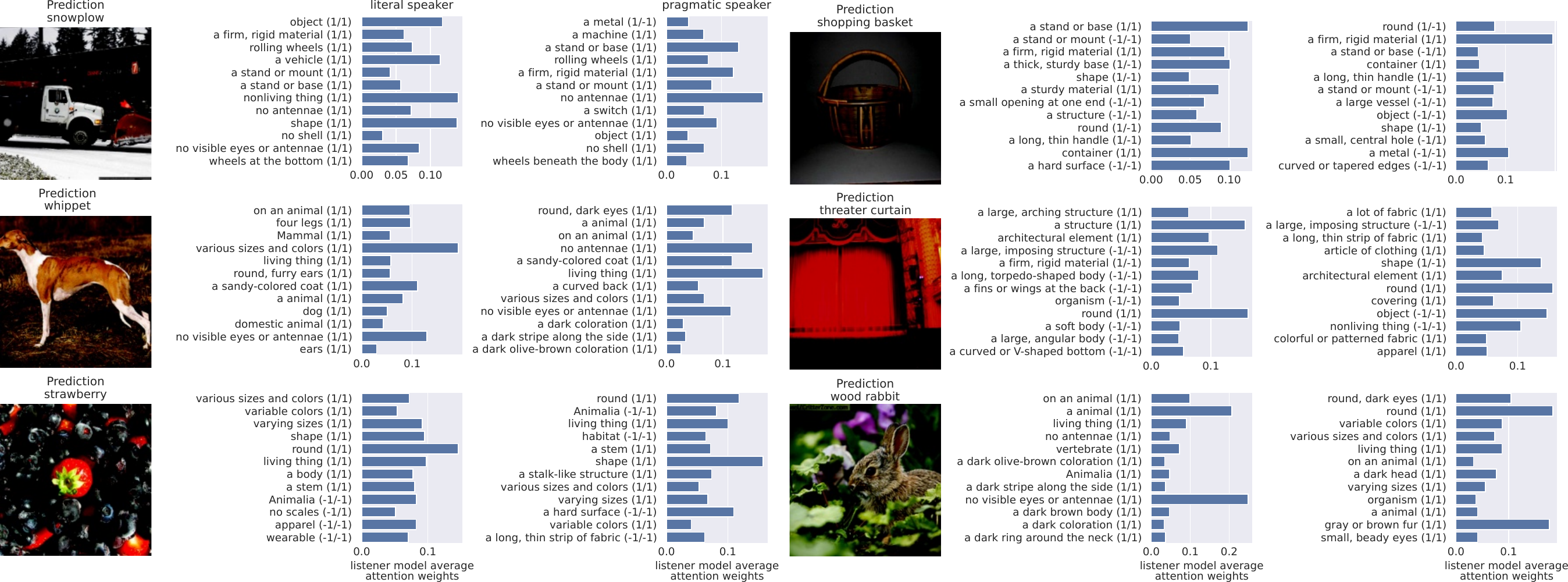}
    \caption{Additional results on the ImageNet dataset for a literal speaker and a pragmatic speaker. We include cases where the classifier and both listeners correctly retrieve the ground-truth label of the image.}
\end{figure}

\newpage
\clearpage
\section{Figure credits}
All diagrams were created with a Lucidchart Educational account (\url{https://lucidchart.com}). The doctor icon is by M. Tohirin from Noun Project (\url{https://thenounproject.com/creator/toelmuhammad/}), CC BY 3.0. The patient icon is by Denicon from Noun Project (\url{https://thenounproject.com/creator/denicon/}), CC BY 3.0. The robot icon is from the NVIDIA Cloud Architecture pack © 2025 NVIDIA Corporation, used with permission.
\end{document}